\documentclass[a4paper,fleqn]{cas-dc}

\usepackage[authoryear]{natbib}
\usepackage{bm}
\usepackage{xcolor}
\usepackage{colortbl}
\usepackage{fontawesome5}
\usepackage{xspace}
\usepackage{mathtools}
\usepackage{graphicx}      
\usepackage{caption}       
\usepackage{booktabs}      
\usepackage{array}         
\usepackage{float}         
\usepackage{multicol}
\usepackage{soul}
\soulregister{\cite}7
\soulregister{\ref}7

\newcommand{\red}[1]{\textcolor{red}{#1}}

\newcommand{\green}[1]{\textcolor{green}{#1}}

\newcommand{\blue}[1]{\textcolor{blue}{#1}}

\definecolor{Gray}{gray}{0.90}
\definecolor{mygray}{rgb}{.80,.80,.80}
\definecolor{myblue}{rgb}{0.94, 0.95, 1.0}
\definecolor{mylightblue}{rgb}{0.96, 0.97, 1.0}
\definecolor{forestgreen}{rgb}{0.0, 0.5, 0.0}
\definecolor{ashgrey}{rgb}{0.7, 0.75, 0.71}
\definecolor{darkorange}{rgb}{1.0, 0.55, 0.0}

\definecolor{backred}{RGB}{255, 190, 190}
\definecolor{backblue}{RGB}{210, 230, 250}
\newcommand{\high}{\cellcolor{backblue}}

\newcommand{\best}{\cellcolor{backred}}

\newcommand{\dataset}{\textsc{AVI-Math}\xspace}

\usepackage{tcolorbox}
\definecolor{myboxcolor}{RGB}{255,255,255} 
\definecolor{myframe}{RGB}{0,0,128} 

\newtcolorbox{mybody}{
  colback=myboxcolor,
  colframe=myframe,
  boxrule=1pt, 
  left=1pt,
  right=1pt,
  top=1pt,
  bottom=1pt,
}

\newcommand{\icoyes}{\textcolor{forestgreen}{\faIcon{check-circle}}\xspace}

\newcommand{\icono}{\textcolor{ashgrey}{\faIcon{times-circle}}\xspace}

\begin{document}
\let\WriteBookmarks\relax
\def\floatpagepagefraction{1}
\def\textpagefraction{.001}

\shorttitle{UAV-Math}    

\shortauthors{Yue Zhou}  

\title [mode = title]{Multimodal Mathematical Reasoning Embedded in Aerial Vehicle Imagery: Benchmarking, Analysis, and Exploration} 



%

\author[1,2]{Yue Zhou}[orcid=0000-0002-3080-6721]

\cormark[1]


\ead{buptzhouyue@gmail.com}



\author[3]{Litong Feng}
\author[2]{Mengcheng Lan}
\author[4]{Xue Yang}
\author[5]{Qingyun Li}
\author[2]{Yiping Ke}
\author[4]{Xue Jiang}
\author[3]{Wayne Zhang}


\affiliation[1]{organization={East China Normal University},
            city={Shanghai},
            postcode={200241}, 
            country={China}}

\affiliation[2]{organization={Nanyang Technological University},
            city={Singapore},
            postcode={639798}, 
            country={Singapore}}
            
\affiliation[3]{organization={SenseTime Research},
            city={Hong Kong},
            country={China}}

\affiliation[4]{organization={Shanghai Jiao Tong University},
            city={Shanghai},
            postcode={200240}, 
            country={China}}
            
\affiliation[5]{organization={Harbin Institute of Technology},
            city={Harbin},
            postcode={150006}, 
            country={China}}

\cortext[1]{Corresponding author.}

\fntext[1]{This work was partially conducted at Nanyang Technological University. This work was supported by National Natural Science Foundation of China (62506229), and Natural Science Foundation of Shanghai (25ZR1402268).}


\begin{abstract}
Mathematical reasoning is critical for tasks such as precise distance and area computations, trajectory estimations, and spatial analysis in unmanned aerial vehicle (UAV) based remote sensing, yet current vision-language models (VLMs) have not been adequately tested in this domain. To address this gap, we introduce \dataset, the first benchmark to rigorously evaluate multimodal mathematical reasoning in aerial vehicle imagery, moving beyond simple counting tasks to include domain-specific knowledge in areas such as geometry, logic, and algebra. The dataset comprises 3,773 high-quality vehicle-related questions captured from UAV views, covering 6 mathematical subjects and 20 topics. The data, collected at varying altitudes and from multiple UAV angles, reflects real-world UAV scenarios, ensuring the diversity and complexity of the constructed mathematical problems. In this paper, we benchmark 14 prominent VLMs through a comprehensive evaluation and demonstrate that, despite their success on previous multimodal benchmarks, these models struggle with the reasoning tasks in \dataset. Our detailed analysis highlights significant limitations in the mathematical reasoning capabilities of current VLMs and suggests avenues for future research. Furthermore, we explore the use of Chain-of-Thought prompting and fine-tuning techniques, which show promise in addressing the reasoning challenges in \dataset. Our findings not only expose the limitations of VLMs in mathematical reasoning but also offer valuable insights for advancing UAV-based trustworthy VLMs in real-world applications. The code, and datasets will be released at \href{https://github.com/VisionXLab/avi-math}{https://github.com/VisionXLab/avi-math}.
\end{abstract}





\maketitle

\section{Introduction}

Mathematical reasoning plays a pivotal role in unmanned aerial vehicle (UAV) based remote sensing tasks such as precise distance measurement, area computation, trajectory estimation, and spatial analysis~\citep{alvarez2021uav,banerjee2020time}. These tasks often require accurate and interpretable reasoning. While deep learning has achieved significant success in remote sensing (RS), its black-box nature raises concerns regarding transparency and reliability, particularly in high-stakes applications~\citep{hohl2024opening}. The advent of vision-language models (VLM)~\citep{yin2023survey}, which exhibit strong mathematical reasoning capabilities, offers a new approach to developing reliable autonomous UAV systems~\citep{wang2024trustworthy}. Unlike traditional deep learning models, VLMs can offer a transparent reasoning process. 
To ensure the development of trustworthy UAV-based VLMs, it is crucial to rigorously assess their multimodal mathematical reasoning abilities.

Several RS Visual Question Answering (VQA) datasets \citep{lobry2020rsvqa,zheng2021mutual,zhang2023crsvqa} have been developed to evaluate multimodal question answering systems. However, these datasets predominantly focus on visual perception tasks, with only a limited number of math-related questions. Furthermore, these math questions tend to be overly simplistic, often restricted to tasks like object counting or basic 2D spatial reasoning, which do not reflect the complexity of real-world RS applications that require domain-specific knowledge in areas like geometry, logic, and algebra. This lack of comprehensive evaluation for mathematical reasoning leaves a critical gap in assessing the true capabilities of current VLMs.

To address this gap, we introduce \dataset, the first comprehensive multimodal benchmark specifically designed to evaluate mathematical reasoning in Aerial Vehicle Imagery (AVI). Unlike existing datasets, \dataset moves beyond simple counting tasks and incorporates domain-specific knowledge across six mathematical subjects — geometry, logic, statistics, arithmetic, counting, and algebra — ensuring a more rigorous and realistic assessment of VLMs' capabilities. The benchmark comprises 3,773 high-quality vehicle-related questions, captured from UAV imagery across 11 distinct 4K resolution scenes. These scenes vary in terms of above-ground level (AGL) and pitch angles, simulating real-world UAV flight conditions. This diversity ensures that the data reflects the complexity of real-world UAV operations, challenging models to reason about mathematical problems under diverse and dynamic conditions.

Given the complexity of the mathematical reasoning tasks in \dataset, we explore methods to enhance VLM performance. We introduce a new instruction fine-tuning dataset with 215k samples and employ LoRA~\citep{hu2021lora} to enable models to better integrate domain-specific knowledge. Furthermore, we investigate several Chain-of-Thought prompting techniques~\citep{wei2022chain,wang2023plan,wu2023despcot,mitra2024ccot}, which encourage models to adopt a multi-step reasoning approach, more closely mimicking human cognitive processes.

This work selects vehicles as the main subject and provides a preliminary exploration of mathematical problems in UAV imagery, without involving multisource RS images or complex sensor characteristics. As illustrated in Tab. \ref{tab:geomath}, GPT-4o demonstrates superior performance in five subjects. However, even the highest overall accuracy achieved is only 34.6\%.  We highlight the challenges that high-resolution UAV images pose to VLMs. We hope that \dataset will serve as a valuable resource, providing a benchmark for the future development of trustworthy multimodal UAV interpretation systems.

In summary, our key contributions are as follows:

\begin{itemize}
  \item We introduce \dataset, the first multimodal benchmark for mathematical reasoning in UAV imagery, covering six subjects and real-world UAV scenarios.
  
  \item We provide a comprehensive analysis, uncovering the limitations of current VLMs in mathematical reasoning and offering insights for future improvements.
  
  \item We explore the potential of Chain-of-Thought prompting and fine-tuning techniques to enhance VLM performance, providing a 215k-sample instruction set for VLMs to learn domain-specific knowledge in UAV scenarios.
  
\end{itemize}

\section{Related Work}

\begin{table}
    \caption{Comparison between existing RS VQA datasets and our fine-tuning instruction set, \textsc{AVI-Math-215k}.}
\resizebox{\columnwidth}{!}{
  \centering
  \setlength{\tabcolsep}{1pt}
  \begin{tabular}{lcccc}
    \toprule
    Dataset & Image Width & \#Questions & \#Subjects & Rationale \\
    \midrule
    RSVQA-LR\citep{lobry2020rsvqa} & 512 & 77,232  & 2 & \icono  \\
    RSVQA-HR\citep{lobry2020rsvqa} & 512 & 1,066,316  &  2 & \icono  \\
    RSVQAxBEN\citep{lobry2021rsvqaxben} & 20$\sim$120 & 14,758,150  & 0  & \icono  \\
    FloodNet \citep{maryam2021floodnet} & 4,000 &  11,000 & 1  & \icono  \\
    RSIVQA \citep{zheng2021mutual} & 256$\sim$4,000 &  111,134 &  2 & \icono  \\
    CDVQA \citep{yuan2022cdvqa} &512  & 122,000  &  1 & \icono  \\
    VQA-TextRS \citep{al2022open} & 256$\sim$600 & 6,245  &  1 & \icono  \\
    CRSVQA \citep{zhang2023crsvqa} & 600 & 4,644  & 2  & \icono  \\
    RSIEval \citep{hu2023rsgpt}& 512 & 936 &  2 & \icono  \\
    EarthVQA \citep{wang2024earthvqa} & 1,024 & 208,593  &  2 & \icono  \\
    VRSBench \citep{li2024vrsbench} & 512 &  123,221 &  2 & \icono  \\
    \hline
    \textsc{AVI-Math-215k} & 4,000 & 215,658  & \textbf{6}  & \icoyes  \\
    \bottomrule
  \end{tabular}}
    \label{tab:vqa_dataset}
\end{table}

\begin{figure}
 \centering
 \vspace{-1mm}
\includegraphics[width=0.8\linewidth]{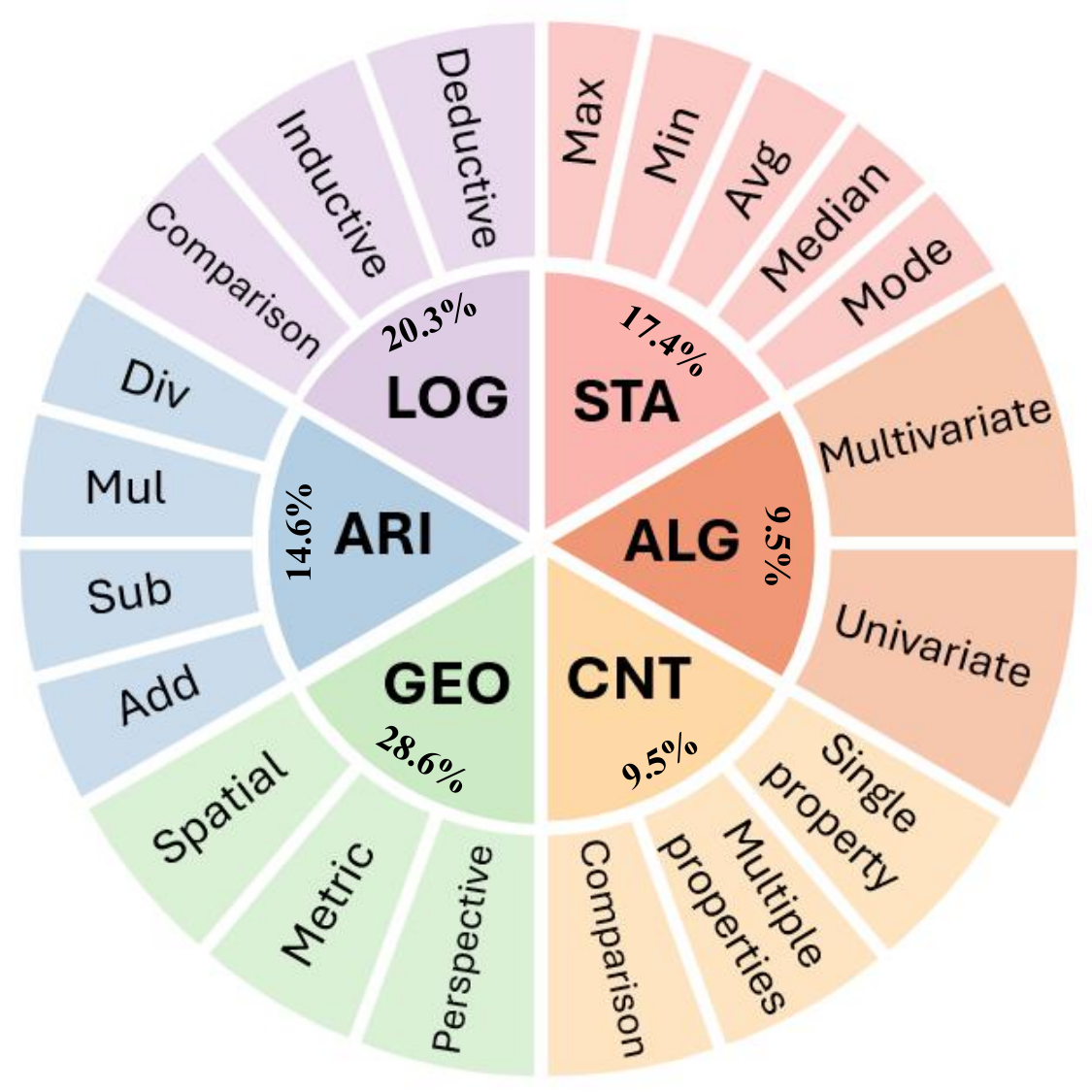}
 \caption{Question types covered by \dataset. ARI: arithmetic, CNT: counting, ALG: algebra, STA: statistics, LOG: logic, GEO: geometry. }
 \label{fig:source_dataset}
\end{figure}

Several benchmarks~\citep{lu2024mathvista,wang2024mathvision,liu2024cmmmath} have been proposed to evaluate the multimodal mathematical reasoning capabilities of VLMs. However, these benchmarks primarily focus on abstract mathematical theory and computation, often relying on simple visual contexts like small figures, charts, or a few natural images. This limits their ability to test models in real-world, domain-specific settings such as RS, where understanding complex spatial relationships and geometric transformations is critical. Unlike these existing benchmarks, our work introduces a domain-specific multimodal mathematical reasoning benchmark that leverages high-resolution UAV images, providing a more realistic and challenging visual context for evaluating the mathematical reasoning abilities of VLMs.

Current RS VQA datasets primarily focus on single-step reasoning, such as land cover classification or building identification~\citep{lobry2020rsvqa}, which oversimplifies the complex reasoning processes often required in real-world RS applications. Furthermore, existing vision-language geo-foundation model (VLGFM) benchmarks~\citep{hu2023rsgpt,li2024vrsbench} typically provide only the final answer, without capturing intermediate reasoning steps, limiting transparency and preventing a thorough evaluation of reasoning validity—particularly crucial in high-stakes scenarios. While RS VQA datasets~\citep{lobry2020rsvqa,zheng2021mutual,al2022open} have advanced the application of VQA systems in UAV-based tasks such as disaster assessment~\citep{maryam2021floodnet}, change detection~\citep{yuan2022cdvqa}, and urban planning~\citep{wang2024earthvqa}, they often involve only simple math problems. To address these limitations, we present a mathematical VQA instruction dataset, \textsc{UAV-Math-215k}. As shown in Tab. ~\ref{tab:vqa_dataset}, It not only covers six mathematical subjects but also incorporates multistep reasoning, with reasoning steps ranging from 2 to 6. This paper presents six types of mathematical problems, as illustrated in the Fig. ~\ref{fig:source_dataset}, thereby extending the boundaries of existing remote sensing VQA.

\section{The \dataset Dataset}
\label{sec:dataset}

As mentioned above, there is a noticeable gap in the RS VQA benchmarks, which mainly focus on evaluating the perceptual capabilities of models while neglecting the mathematical capabilities. Therefore, our dataset, \dataset, aims to bridge this gap by providing a robust evaluation benchmark for mathematical reasoning intertwined with RS visual perception. In this section, we present the \dataset, following the steps of data collection, metadata annotation, question design, and question generation. Finally, we perform data analysis on the dataset.



\subsection{Data Collection}

To the best of our knowledge, there is currently no dedicated mathematical dataset specifically designed for remote sensing. Existing open-source RS datasets~\citep{xia2018dota,li2020object} often lack sensor metadata and provide limited target attributes. Consequently, these datasets can only support the formulation of simple mathematical problems, such as counting the object according to its color or judging the relative position in the image. To develop a more specialized and diverse mathematical dataset, we use UAVs to collect data from scratch. This approach ensures comprehensive access to sensor parameters and detailed information about ground targets. To enhance the diversity of mathematical problems, we choose vehicles as the subject of drone photography. Compared to buildings or land cover~\citep{yang2010bag}, vehicles have richer attributes and more fine-grained categories. Data collection is divided into two parts: aerial imagery and ground video.

\paragraph{Aerial Imagery} 
All aerial images in \dataset were collected with a small UAV platform, DJI Mini3. The dataset consists of 4K high-resolution RS images from 11 distinct scenes, captured at 9 different above-ground levels (AGLs) and 3 pitch angles. This implies that these RS images possess different spatial resolutions and perspectives. In addition, the collected images cover a variety of weather scenarios, such as sunny, cloudy, and rainy days, along with different lighting conditions.

\paragraph{Ground Video} 
We record ground videos from the same areas to facilitate accurate annotation of vehicle brands and models. Specifically, we select time slots with relatively low vehicular mobility, avoiding rush hours and meal times. Additionally, to mitigate the vehicle mismatch between drone images and ground videos caused by vehicle entry and exit, we capture two sets of ground videos before and after the drone captures aerial photos. This ensures that vehicles entering or exiting the scene halfway through the capture are recorded in the videos. However, there are instances where vehicles pass through the scene briefly, leading to cases where they are not captured in either video. In such situations, we mask these vehicles with black masks. Our statistical analysis revealed that mask occlusion vehicles account for only 6.92\% of the total, representing a relatively small proportion.



\begin{table}
\caption{Details of metadata, where most vehicle attributes are obtained from the ground video.}
\centering
\small
\renewcommand\tabcolsep{2.0pt} 
\begin{tabular}{cp{5cm}}
    \toprule
    \textbf{Type} & \textbf{Details} \\
    \midrule
    \multirow{2}{*}{\parbox{3cm}{\centering Camera \\ \centering parameters}}
    & Focal length, ISO, pixel size, shutter speed, aperture, sensor size, image resolution, pitch angle, AGL, latitude, longitude, timestamp. \\
    \midrule
    \multirow{2}{*}{\parbox{3cm}{\centering Vehicle \\ \centering attributes}}
    & Location of pixel coordinate system, rotated bounding box, front direction, brand, model, color, type, powertrain, length, width, height, sunroof, roof rack, max price, min price, number of doors / seats. \\
    \bottomrule
\end{tabular}

\label{tab:mata_data}
\end{table}

\subsection{Metadata Annotation}

The metadata can be categorized into two main components. The first includes camera-related parameters, such as intrinsic parameters (focal length, pixel size, sensor dimensions) and extrinsic parameters (pitch angle, AGL). These are extracted from the raw data from the drone. The second component pertains to vehicle fine-grained attributes, which require manual annotation. To accurately describe the length and width of vehicles, we use rotated bounding boxes to annotate their positions~\citep{yang2022detecting}. Then, a 360-degree angle representation is used to depict the vehicle's orientation~\citep{hu2023trr360d}. Identifying specific vehicle brands from aerial imagery presents a significant challenge for human annotators, and as a result, existing publicly available RS vehicle datasets have not achieved brand-level annotations~\citep{mundhenk2016large,zhu2021detection}.

However, leveraging the previously mentioned ground videos, we successfully created the first RS vehicle dataset with fine-grained attributes, identifying vehicles down to the model level within each brand. Specifically, we match the vehicles in the aerial image with the vehicles in the ground video one by one according to their locations and then call the DCD's API \footnote{\url{https://dcdapp.com}} to identify the specific model based on the vehicle's appearance and logo in the ground image. For vehicles whose models could not be identified, we used a black mask to cover them from the image. Then we used the DCD car database to obtain detailed attributes, such as the size and price of each car. Tab. \ref{tab:mata_data} presents all the meta-information collected in our dataset.

The annotation quality in this study is ensured for the following reasons. First, we only need to annotate the vehicle's brand and model, while fine-grained attributes like price and powertrain can be accurately retrieved from the vehicle parameters table. Second, although the dataset contains over 16k vehicle samples from drone images, there are only 814 unique vehicle instances. We manually annotate these instances using ground-level videos and map them to aerial images. Finally, we designed a rigorous annotation pipeline: 1) Call the API to predict the brand and model from front/rear views.
2) Call the API to predict the brand from cropped logos.
3) The annotators manually verify the brand and model.
4) For inconsistent predictions, annotators review additional perspectives from ground-level videos to make a judgment based on the badge located on the front or rear of the vehicle.
5) Discard samples if remain uncertain.


To more objectively assess annotation quality, we followed mainstream evaluation methodologies ~\citep{wang2024novelqa,sandmann2025benchmark} by re-annotating 10\% of vehicle samples and conducting cross-validation using the multi-class Cohen's Kappa coefficient. The achieved average consistency score of 0.9768 indicates exceptionally high annotation reliability.

\subsection{Coordinate System Transformation}
\label{sec:coord_system_transformation}

To evaluate the spatial perception capability of the VLM, it is necessary to obtain the vehicle's coordinates in three-dimensional space through coordinate transformation. The related camera parameters include pitch angle $\theta$, AGL $H$, focal length $f$, and pixel size $p$. The camera intrinsics are fixed, and photos are captured on windless days with the drone stationary, ensuring stable camera extrinsics and accurate drone altitude. The complete derivation processes for two coordinate system transformations are provided below. 

\begin{figure*}[t!]
  \centering
\includegraphics[width=1.0\textwidth]{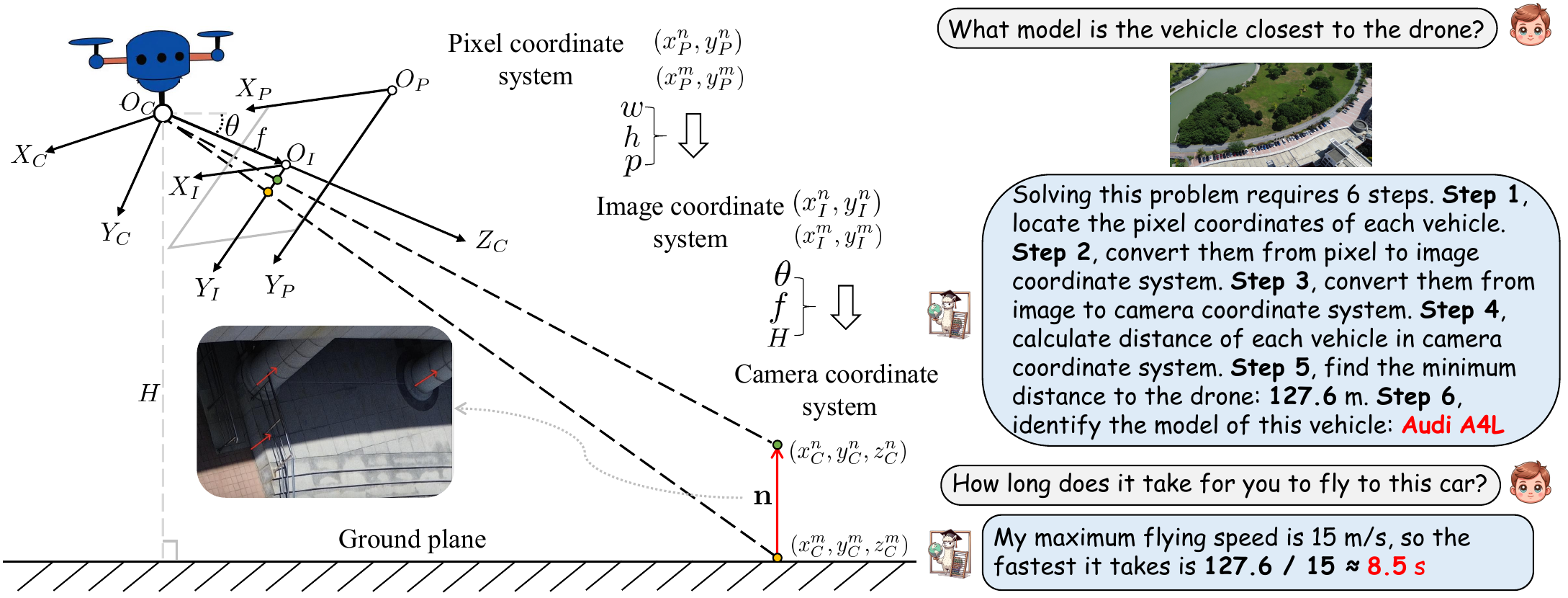}
  \vspace{-5mm}
  \caption{Mathematical modeling of UAV Scenes and examples for geometric question.}
\label{fig:math_modeling}
\end{figure*}

The transformation between the pixel coordinate system and the image coordinate system can be represented by an affine matrix, as follows:
\begin{equation}
\begin{split}
\left[ \begin{array}{c}
	x_P\\
	y_P\\
	1\\
\end{array} \right] =\left[ \begin{matrix}
	\frac{1}{p}&		0&		\frac{w}{2}\\
	0&		\frac{1}{p}&		\frac{h}{2}\\
	0&		0&		1\\
\end{matrix} \right] \left[ \begin{array}{c}
	x_I\\
	y_I\\
	1\\
\end{array} \right] 
\end{split}
\label{Equ:1}
\end{equation}
where $p$ represents the pixel size of sensor. $\frac{w}{2}$ and $\frac{h}{2}$ denote the origin offsets, with the origin of the pixel coordinate system typically located at the image's top-left corner. Given the pixel coordinates $(x_P, y_P)$ of a certain point, its corresponding image coordinates $(x_I, y_I)$ can be calculated as follows:
\begin{equation}
\begin{split}
\begin{cases}
x_I = (x_P - w/2) \cdot p \\
y_I = (y_P - h/2) \cdot p
\end{cases}
\end{split}
\label{Equ:1}
\end{equation}

The transformation from the camera coordinate system to the image coordinate system is a conversion from three-dimensional to two-dimensional coordinates. Assuming the focal length of the camera is $f$, then we have
\begin{equation}
\begin{split}
z_c\left[ \begin{array}{c}
	x_I\\
	y_I\\
	1\\
\end{array} \right] =\left[ \begin{matrix}
	f&		0&		0&		0\\
	0&		f&		0&		0\\
	0&		0&		1&		0\\
\end{matrix} \right] \left[ \begin{array}{c}
	x_C\\
	y_C\\
	z_C\\
	1\\
\end{array} \right] 
\end{split}
\label{Equ:2}
\end{equation}
where $z_C$ denotes the depth of the point, which can be obtained by a depth camera (binocular or structured light). Because the drone camera we are using cannot provide depth information, we need to find another way.

When the ground satisfies the ground plane assumption, given the AGL of the drone and the pitch angle of the camera, the ground plane equation in the camera coordinate system is as follows:
\begin{equation}
\begin{split}
-\cos{\theta} \cdot Y_C - \sin{\theta} \cdot Z_C + H = 0 
\end{split}
\label{Equ:3}
\end{equation}

The equation of the line connecting the camera origin to the projection point on the pixel plane in the camera coordinate system is given by:
\begin{equation}
\begin{split}
\begin{cases}
X_C = x_I \cdot t \\
Y_C = y_I \cdot t\\
Z_C = f \cdot t
\end{cases}
\end{split}
\label{Equ:4}
\end{equation}
Substituting the line equation into the ground plane equation yields:
\begin{equation}
\begin{split}
t = \frac{H}{y_I\cos{\theta} + f\sin{\theta} }
\end{split}
\label{Equ:5}
\end{equation}
Substituting $t$ back into the line equation yields:
\begin{equation}
(\frac{x_I  H}{y_I\cos{\theta} + f\sin{\theta}}, \frac{y_I  H}{y_I\cos{\theta} + f\sin{\theta}}, \frac{f  H}{y_I\cos{\theta} + f\sin{\theta}})
\label{Equ:5}
\end{equation}

To preserve the spatial mapping between camera coordinates and pixel coordinates, we refrained from cropping the 4K images to increase the dataset size, as is commonly done in most remote sensing datasets.

\begin{table*}[h!]
\caption{Examples of six mathematical reasoning subjects in \dataset. (Part 1)}
\label{tab:math_examples-1}
\centering
\renewcommand\tabcolsep{1.0pt} 
\small
\begin{tabular}{m{1cm}p{14cm}}
    \toprule
    \textbf{Type} & \textbf{Example} \\
    \midrule
    GEO & 
    \begin{minipage}[s][4.7cm]{0.195\linewidth}
    \vspace{0mm}
    \hspace{0mm}
    \includegraphics[height=4.6cm]{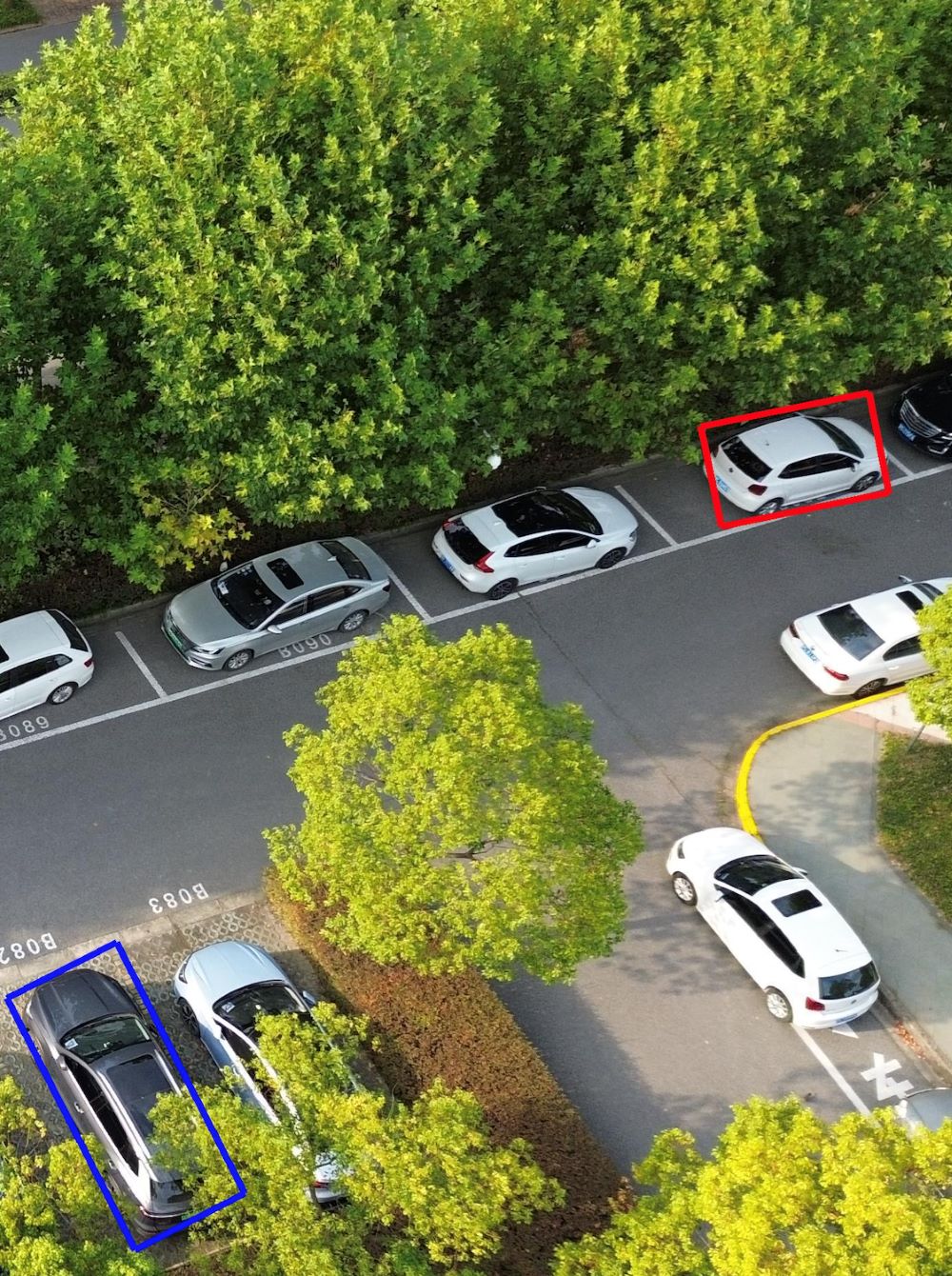}
    \end{minipage}
    \hspace{7mm}
    \begin{minipage}[s][4.7cm]{0.7\linewidth}
    \textbf{Context:} The sensor parameters that may be used are as follows: \\
    Focal Length: 12 millimeters. Pixel Size: 0.004325 millimeters. Image Width: 4000 pixels. Image Height: 2250 pixels.\\
    \textbf{Question:} How many meters are the two vehicles in the red and blue boxes apart?\\
    \textbf{Rationale:} 
    Step 1, locate the center point of two vehicles in the red and blue boxes: [3554, 1051] and [2583, 1974]\\
    Step 2, convert them from pixel to image coordinate system: ... \\
    Step 3, convert them from image to camera coordinate system: ...\\
    Step 4, calculate the distance in the camera coordinate system: $\sqrt{(24.4-6.8)^2 + (-1.2-9.9)^2 + (43.6-32.5)^2} \approx 23$ \\
    \textbf{Answer: 23} 
    \end{minipage}
    \\
    \midrule
    LOG & 
    \begin{minipage}[s][2.45cm]{0.26\linewidth}
    \includegraphics[height=2.45cm]{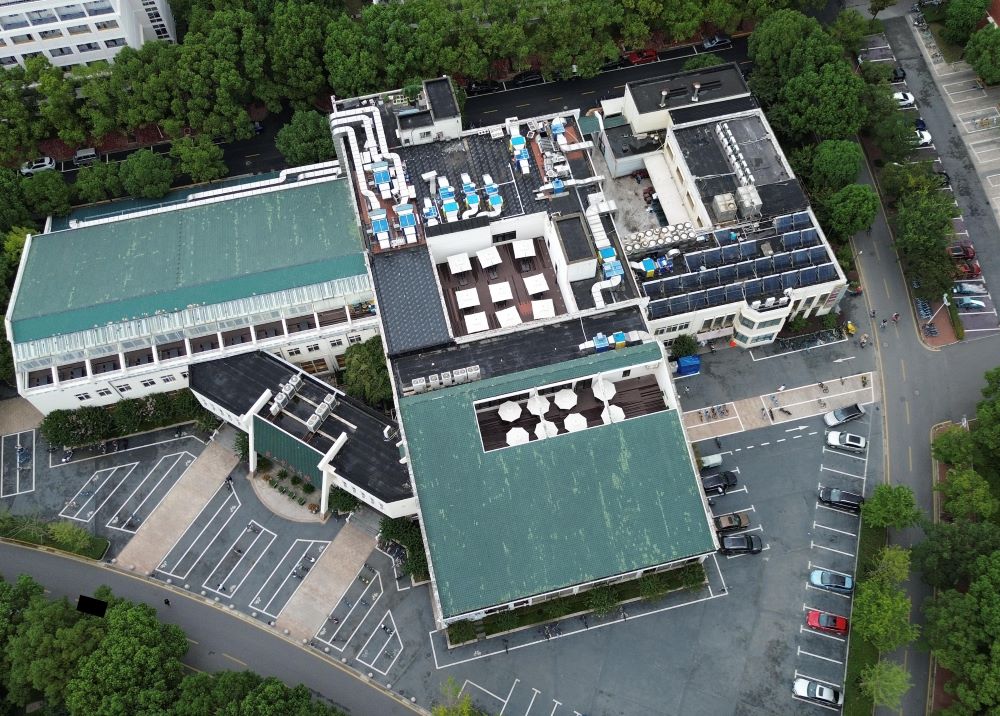}
    \end{minipage}
    \begin{minipage}[s][2.3cm]{0.7\linewidth}
    \textbf{Question:} There were 27 cars in this area an hour ago, how many cars have entered this area at least within an hour? \\
    \textbf{Rationale:}
    Step 1, count all current vehicles: 29 \\
    Step 2, the number of cars entering the area is at least equal to the increase in the number of cars in this area: 29-27=2 \\
    \textbf{Answer: 2} 
    \end{minipage}
    \\
    \midrule
    STA & 
    \begin{minipage}[s][2.2cm]{0.195\linewidth}
    \vspace{0mm}
    \hspace{0mm}
    \includegraphics[height=2.07cm]{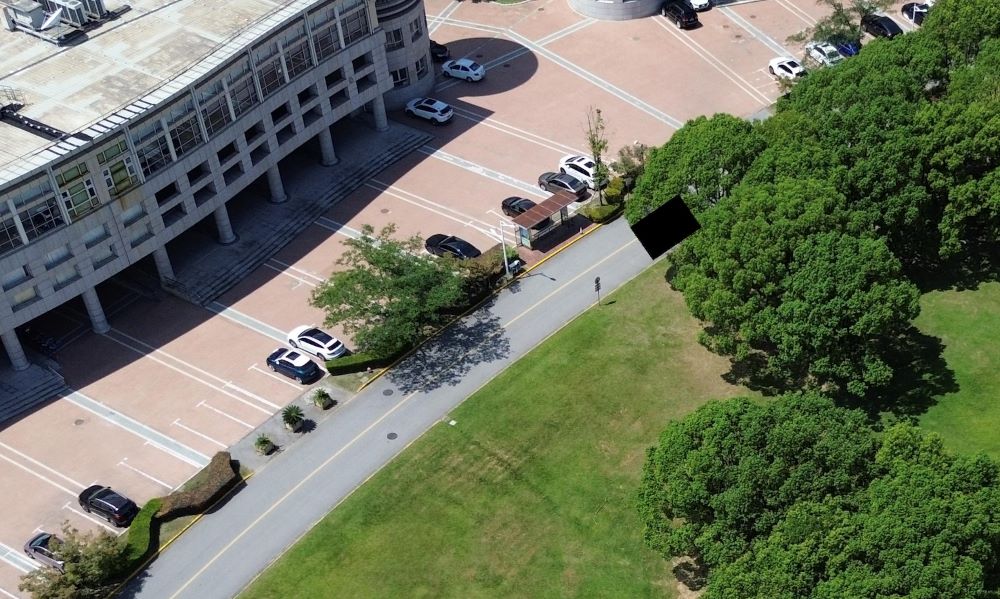}
    \end{minipage}
    \hspace{7mm}
    \begin{minipage}[s][2.2cm]{0.7\linewidth}
    \textbf{Question:} What color of vehicle is most common in the image?\\
    \textbf{Rationale:} 
    Step 1, identify the color of all vehicles: [`white', `brown', ...]\\
    Step 2, count vehicles for each color: \{`white': 7, `brown': 3, ...\}\\
    Step 3, sort to get the most common color: white \\
    \textbf{Answer: white} 
    \end{minipage}
    \\
    \bottomrule
\end{tabular}
\end{table*}

\subsection{Question Design and Generation}

Recent works~\citep{li2024vrsbench,xu2024rs} adapt GPT to automatically generate RS VQAs, to reduce manual labor. Compared to template-based methods, model-generated questions exhibit greater diversity. However, in mathematical benchmarks, the rigor of the questions is paramount. Given the current performance of GPT on multimodal mathematical benchmarks~\citep{lu2024mathvista,wang2024mathvision}, we cannot fully trust it. Therefore, we choose a template-based question generation approach, which offers more control over content compared to generative models. To compensate for the lack of diversity, we designed more than 80 templates based on 20 topics. As there is no established mathematical reasoning benchmark in remote sensing, we reference two general benchmarks\textemdash MathVista~\citep{lu2024mathvista} and MathVision~\citep{wang2024mathvision}\textemdash to categorize our math questions into six types. These benchmarks are widely used by mainstream VLMs and have become standards for multimodel mathematical reasoning. Math problems often test multiple aspects, such as statistics, which inevitably involve basic arithmetic knowledge. Following the priority system of general benchmarks, questions involving mean, sum, or other statistical variables are classified as statistics. Tab. \ref{tab:math_examples-1} and \ref{tab:math_examples-2} show examples of the 6 subjects of questions.

\subsubsection{Geometry} 
The geometric questions extend the spatial relationships in RS VQA from the 2D-pixel plane to the 3D real world. Relevant domain knowledge includes metric geometry and perspective geometry. Fig. \ref{fig:math_modeling} illustrates a typical UAV reconnaissance scenario. Given the relatively flat terrain of the shooting area, we can assume that it satisfies the assumption of a flat surface~\citep{novak2017vehicle}. We validated this assumption by placing normal vectors $\mathbf{n}$ on reference objects such as poles.  Given the camera parameters, the pixel coordinates of a car can be used to compute its camera coordinates.

Based on these, the closest vehicle can be identified and the shortest flight time can be estimated based on the speed of the drone. In addition, we can estimate the area of the captured region as well as the size and orientation of the vehicles.

\begin{table*}[h!]
\caption{Examples of mathematical reasoning subjects in \dataset. (Part 2)}
\label{tab:math_examples-2}
\centering
\renewcommand\tabcolsep{1.0pt} 
\small
\begin{tabular}{m{1cm}p{14cm}}
    \toprule
    \textbf{Type} & \textbf{Example} \\
    \midrule
    ARI & 
    \begin{minipage}[s][3.05cm]{0.195\linewidth}
    \vspace{0mm}
    \hspace{0mm}
    \includegraphics[height=2.9cm]{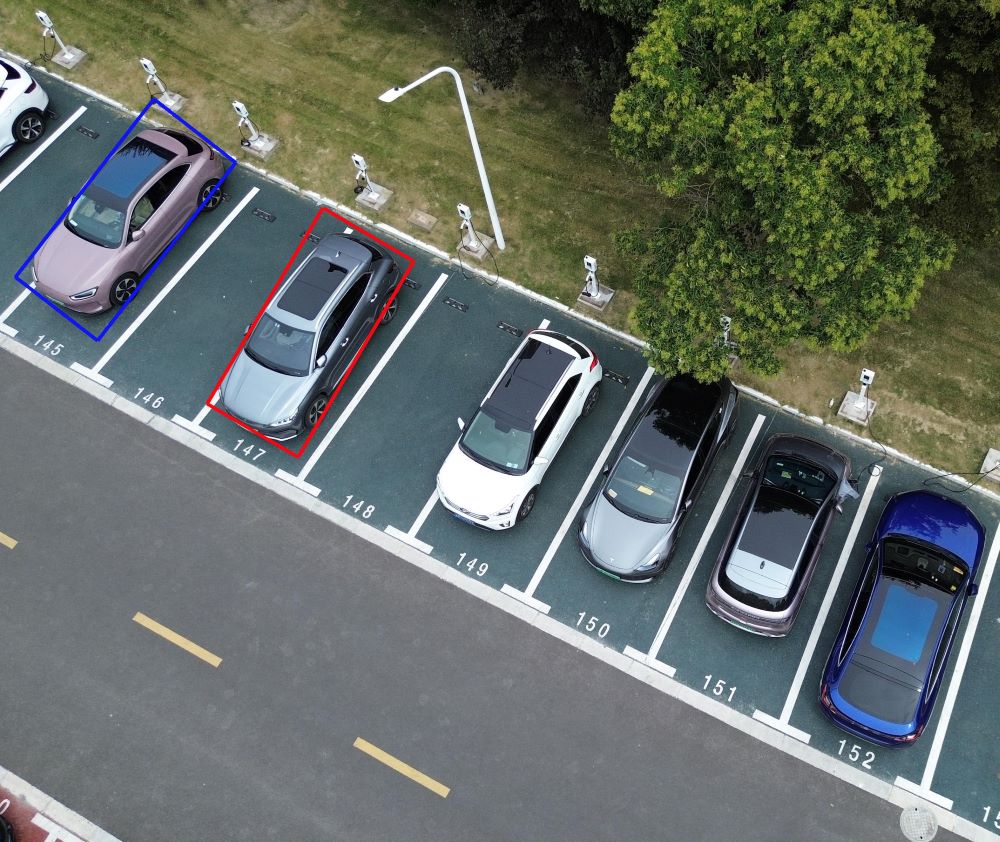}
    \end{minipage}
    \hspace{7mm}
    \begin{minipage}[s][3.35cm]{0.7\linewidth}
    \textbf{Context:} The vehicle price dictionary that may be used is as follows: \{`nio ec6': 385000, `byd dolphin': ...\}\\
    \textbf{Question:} What is the price difference between the car in the red box and the car in the blue car? (Unit: RMB)\\
    \textbf{Rationale:} \\
    Step 1, identify the model of two cars: byd song plus and aito m5\\
    Step 2, query the prices of two vehicles: 155000 and 265000\\
    Step 3, calculate the price difference: 265000-155000=110000 \\
    \textbf{Answer: 110000} 
    \end{minipage}
    \\
    \midrule
    CNT & 
    \begin{minipage}[s][2.2cm]{0.26\linewidth}
    \includegraphics[height=2.15cm]{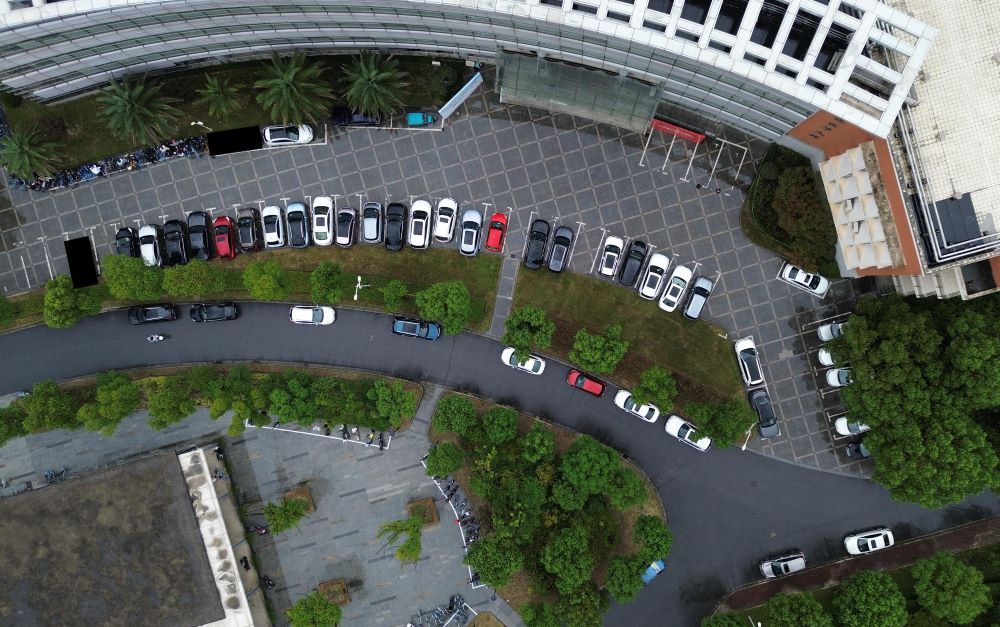}
    \end{minipage}
    \begin{minipage}[s][2.2cm]{0.7\linewidth}
    \vspace{2mm}
    \textbf{Question:} How many SUV vehicles are there in the image?\\
    \textbf{Rationale}: 
    Step 1, identify the type of all vehicles: [`suv', `suv', ...] \\
    Step 2, count all SUV vehicles: 17 \\
    \textbf{Answer: 17} 
    \end{minipage}
    \\
    \midrule
    ALG & 
    \begin{minipage}[s][4.8cm]{0.195\linewidth}
    \vspace{0mm}
    \hspace{0mm}
    \includegraphics[height=4.6cm]{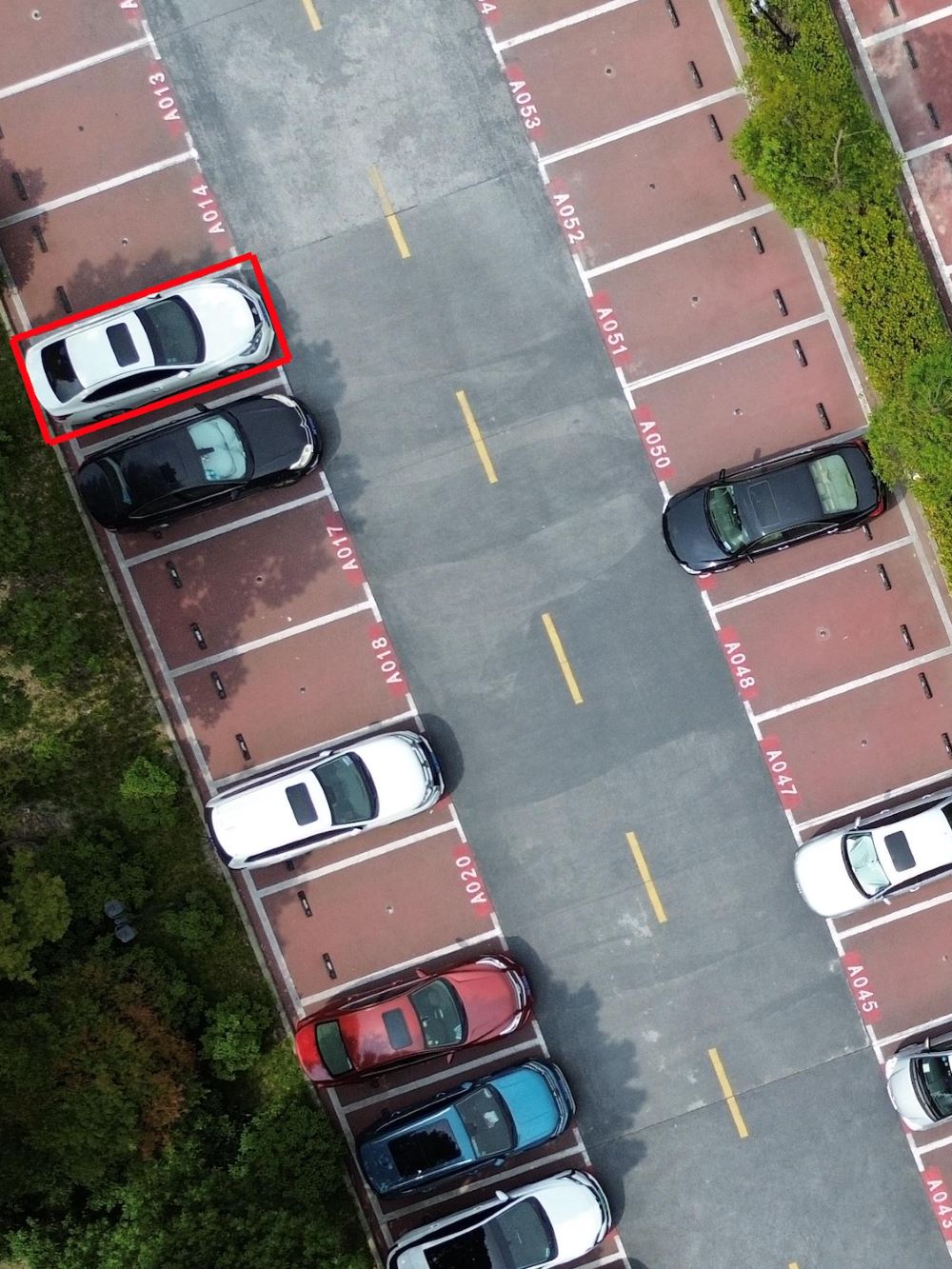}
    \end{minipage}
    \hspace{7mm}
    \begin{minipage}[s][4.85cm]{0.7\linewidth}
    \textbf{Context:} The sensor parameters that may be used are as follows: \\
    Focal Length: 12 millimeters. Pixel Size: 0.004325 millimeters. Image Width: 4000 pixels. Image Height: 2250 pixels.\\
    \textbf{Question:} The equation of the ground plane in the camera coordinate system is: -cos(90)*y-sin(90)*z+40=0. What are the coordinates of the center point of the vehicle in the red box in the camera coordinate system? (Unit: meter)\\
    \textbf{Rationale:} 
    Step 1, locate the center point of the vehicle: [420, 534]\\
    Step 2, convert the center point of the vehicle from the pixel coordinate system to the image coordinate system: [-6, -2]\\
    Step 3, convert the center point of the vehicle from the image coordinate system to the camera coordinate system: [-22, -8, 40]\\
    \textbf{Answer: [-22, -8, 40]} 
    \end{minipage}
    \\
    \bottomrule
\end{tabular}
\end{table*}

\subsubsection{Arithmetic} We construct a series of arithmetic questions, including addition, subtraction, multiplication, and division, based on the prices of the vehicles. For example, questions may ask which of the two cars is more expensive or how many of a certain type of car can be bought with 1 million RMB. Considering that vehicle prices can be unstable due to market fluctuations, we have provided the vehicle models and their corresponding prices in the context field of each problem. We exclude questions that can be answered solely through pure text, ensuring that the model must rely on visual data to arrive at the correct answer. This approach ensures that the model can obtain all the necessary information to solve the current mathematical problems in an offline environment, without the need for RAG~\citep{gao2023retrieval} techniques.

\subsubsection{Counting} By incorporating more fine-grained attributes of the cars, we can construct a wider variety of counting questions with varying levels of difficulty. Related attributes include vehicle types, brands, models, and prices. The generated questions not only involve counting based on single-attribute constraints but also include comparative counting and counting based on multiple-attribute constraints. For example, questions may ask for the number of cars priced above 100,000 RMB or the number of white SUVs. In \dataset, each image contains an average of 25.8 cars. The differences between vehicles are smaller compared to those between different object categories, making the task more challenging.

\subsubsection{Algebra} The algebraic questions are primarily divided into two categories: single-variable algebra and multi-variable algebra. The model needs to use its visual perception capabilities to obtain certain variables and then solve equations to determine the target variable. The relevant domain knowledge includes spatial coordinate system transformations, such as determining the coordinates of a vehicle in the image or camera coordinate system based on its pixel coordinates obtained from the image. We also construct algebraic questions related to prices, such as calculating the price of the vehicle closer to the image bottom based on the total price of two cars and their price ratio.

\subsubsection{Logic} In the design of logic problems, beyond incorporating image-based information, some common-sense knowledge from daily life is introduced. For example, please identify the vehicles in the image that need to visit gas stations regularly. Based on common sense, vehicles that need to visit gas stations are those powered by fuel. Therefore, it can be inferred that the actual intention of the question is to identify the fuel-powered cars.


\subsubsection{Statistics} We design statistical questions based on vehicle prices and sizes, covering maximum, minimum, mean, and mode. Related domain knowledge is metric geometry.

During the question generation phase, we prioritize the selection of images in which vehicles are not significantly occluded by buildings or trees to build the benchmark. The process consists of three steps: 1) generating image-level questions without modifying the images; 2) generating single-instance questions by randomly selecting a vehicle and drawing a rotated bounding box around it as a visual prompt; and 3) generating two-instance questions by randomly selecting two vehicles and drawing their rotated bounding boxes in different colors (e.g. red and blue). Vehicles near the edge of the image are excluded to avoid difficulties due to incomplete visual information. Finally, the generated questions are manually reviewed for accuracy.

\subsection{Data Analysis}

The main statistics of \dataset are presented in Tab \ref{tab:statistics1}. There are three types of questions: multiple choice, free-form, and true or false. The answers to free-form questions are classified as integers, floating numbers, lists, or strings. Variations in pitch angle and AGL ensure the diversity of observation patterns in \dataset. Fig. \ref{fig:source_dataset} and Fig. \ref{fig:math_modeling} show that There are 6 subjects and 20 topics in our benchmark. For each question, we provide a complete reasoning process. The rationale can help VLMs learn to solve math problems through reasoning rather than rote memorization.

\begin{table}
  \caption{Key statistics of \dataset.}
 \centering
 \fontsize{8.2pt}{\baselineskip}\selectfont 
 \renewcommand\tabcolsep{1.0pt} 
 \renewcommand\arraystretch{0.8} 
 \begin{tabular}{lc}
 \toprule
 \textbf{Statistic} & \textbf{Number} \\
 \midrule
  Total questions & 3,773 \\
  ~- Multiple-choice questions & 1,352 (35.8\%) \\
  ~- Free-form questions & 2,181 (57.8\%) \\
  ~- True/False questions & 240 (6.4\%) \\
 \midrule
 Unique number of images & 360 \\
  ~- Pitch angle: 90 & 117 (32.5\%) \\
  ~- Pitch angle: 60 & 126 (35\%) \\
  ~- Pitch angle: 45 & 117 (32.5\%) \\
  ~- Above ground level: low & 138 (38.3\%) \\
  ~- Above ground level: medium & 108 (30.0\%) \\
  ~- Above ground level: high & 114 (31.7\%) \\
 Unique number of questions & 424 \\
 Unique number of answers & 686 \\
 \midrule
 Maximum question length & 236 \\
 Minimum question length & 45 \\
 Average question length & 101.5 \\
 \midrule
 Maximum reasoning steps & 6 \\
 Minimum reasoning steps & 2 \\
 Average reasoning steps & 3.34 \\
 \bottomrule
 \end{tabular}
 \label{tab:statistics1}
 \end{table}

\begin{table}[t!]
\caption{Summary of the 20 different topics in \dataset.}
\centering
\small
\renewcommand\tabcolsep{2.0pt} 
\resizebox{1.0\linewidth}{!}{
\begin{tabular}{lccc}
    \toprule
    \textbf{Topic} & \textbf{Subject} & \textbf{Visual Skill} & \textbf{Application} \\
    \midrule
    Perspective Geometry & GEO & Location & Surveying\\
    Metric Geometry & GEO & Location & Surveying \& Military \\
    Spatial Relation & GEO & FG Recognition, Location & Surveying \& Military \\
    \midrule
    Comparison & LOG & FG Recognition, Visual Prompt & Entertainment\\
    Deduction & LOG & FG Recognition, Visual Prompt & Surveillance\\
    Induction & LOG & FG Recognition & Surveillance \\
    \midrule
    Maximum & STA & FG Recognition, Location & Market Research \\
    Minimum & STA & FG Recognition, Location & Market Research\\
    Mean & STA & FG Recognition, Location & Market Research\\
    Median & STA & FG Recognition, Location & Market Research\\
    Mode & STA & FG Recognition, Location & Market Research\\
    \midrule
    Addition & ARI & FG Recognition & Market Research\\
    Subtraction & ARI & FG Recognition, Visual Prompt & Market Research\\
    Multiplication & ARI & FG Recognition, Visual Prompt & Market Research\\
    Division & ARI & FG Recognition, Visual Prompt & Market Research\\
    \midrule
    Counting (single property) & CNT & FG Recognition & Market Research\\
    Counting (multiple property) & CNT & FG Recognition & Market Research\\
    Counting (comparison) & CNT & FG Recognition & Market Research\\
    \midrule
    Univariate Equation & ALG & Location, Visual Prompt & Surveying \\
    Multivariate Equations & ALG & Location, Visual Prompt & Surveying\\
    \bottomrule
\end{tabular}
}

\label{tab:topic_sum}
\end{table}

Tab. \ref{tab:topic_sum}  describes the visual abilities required for each topic and their application scenarios. Location represents the ability to provide the pixel coordinates of key points. FG recognition,  short for fine-grained recognition, refers to the ability to identify critical visual cues in RS images, including the specific properties and models of vehicles. Visual prompt indicates the capability to determine the referenced target based on various colored boxes added to the image. Surveying suggests that remote sensing professionals can leverage this capability to enhance the efficiency of geological surveys and obtain interpretable and reliable results. Military indicates that it can be used in unmanned warfare to improve the intelligence level of drones. Entertainment indicates that users can utilize this capability to satisfy their curiosity. Surveillance indicates that this capability can be used to monitor activities within a specific area. Market research indicates that automotive companies can leverage this capability to conduct fine-grained analysis of customer preferences within a specific region.



\subsection{AVI-MATH-215K}

To perform parameter-efficient fine-tuning for \textsc{AVI-Math}, we introduce \textsc{AVI-Math-215k}, a multimodal instruction set comprising 215k image-question-answer-rationare samples. The construction process of AVI-MATH-215K aligns with the AVI-MATH Benchmark, utilizing the same batch of collected UAV images and adhering to identical manual data annotation procedures to ensure question quality. The images come from the collected data, excluding the benchmark. We deliberately designed the instruction set without multiple-choice questions, as they may encourage models to exploit shortcuts rather than develop a true understanding of the content. The average length of questions in this instruction set is 105.95 words, with a maximum length of 376 words and a minimum length of 64 words. The average reasoning length is 62.79 words, with a maximum of 1,054 words and a minimum of 24 words. Additionally, we analyzed the number of vehicles involved in the questions. The proportions of questions involving all vehicles in the scene, one vehicle, and two vehicles are 7.26\%, 34.00\%, and 58.74\%, respectively. For questions related to surveying/mapping or pricing, we provide context such as camera/drone parameters or vehicle price lists to assist the model in decision-making. Among the samples, 50.43\% include such contextual information.

\begin{table*}[t!]
    \caption{Accuracy scores on the \dataset. AVG: average accuracy of the six subjects. FRE: free-form question, CHO: multiple choice question, T/F: true or false question. The highest scores among models in each part and overall are highlighted in \blue{blue} and \red{red}. The table exclusively employs the original model weights without fine-tuning.}
\centering
 \small
 \renewcommand\tabcolsep{2.5pt} 
 \renewcommand\arraystretch{1.1} 
 \resizebox{1.0\linewidth}{!}{
    \begin{tabular}{l|l|c|cccccc|ccc|ccc|ccc}
    \toprule
    \multirow{2}{*}{Model} &  
    \multirow{2}{*}{LLM} & 
    \multirow{2}{*}{AVG} & 
    \multicolumn{6}{c}{Subject} &
    \multicolumn{3}{c}{AGL} &
    \multicolumn{3}{c}{Pitch Angle} &
    \multicolumn{3}{c}{Type}\\
    \cline{4-18} 
    & &  & ALG & ARI & CNT & GEO & LOG  & STA & Low & Med & High & 45 & 60 & 90 & FRE & CHO & T/F \\
    \midrule
    Random chance & - & 11.7 & 11.5 & 8.4 & 10.6 & 8.0 & 22.6 & 9.2 & 10.7 & 13.2 & 12.3 & 11.1 & 12.5 & 12.2 & 0.0 & 24.3 & 51.3  \\
    \midrule
    \multicolumn{17}{l}{\hfill \textit{Small-scale VLMs (LLM's Parameters $<$ 10 Billion)} } \\
    \midrule
    GeoChat~\citep{kuckreja2024geochat} & Vicuna-7B & 12.6 & 15.4 & 5.4 & 16.7 & 9.4 & 24.1 & 4.5 & 13.9 & 12.7 & 10.7 & 11.5 & 12.3 & 13.7 & 4.6 & 19.9 & 42.1  \\
    XComposer2~\citep{zhang2023internlm} & InternLM2-7B & 13.6 & 4.0 & 2.9 & 20.0 & 2.3 & 28.7 & 23.4 & 15.4 & 13.5 & 12.0 & 12.0 & 14.4 & 14.7 & 11.0 & 11.8 & 49.6  \\
    Qwen-VL-Chat~\citep{bai2023qwen} & Qwen-7B & 16.5 & 6.7 & 10.7 & 15.8 & 11.4 & 30.2 & 24.1 & 18.8 & 18.1 & 17.8 & 18.1 & 18.4 & 18.3 & 9.5 & 26.9 & 49.2  \\
    LLaVA-v1.5-7B~\citep{liu2023llava} & Vicuna-7B & 18.3 & 12.5 & 11.6 & 15.8 & 12.2 & 35.8 & 22.2 & 19.1 & 22.7 & 18.3 & 19.9 & 18.5 & 21.4 & 10.9 & 28.3 & 54.2  \\
    XComposer2.5~\citep{zhang2023internlm} & InternLM2-7B & 18.5 & 4.0 & 16.4 & 26.1 & 7.2 & 26.8 & 30.3 & 20.4 & 18.6 & 17.6 & 18.4 & 20.0 & 18.5 & 7.7 & 30.8 & 55.0  \\
    DeepSeek-VL~\citep{lu2024deepseek} & DeepSeek-7B-Base & 18.5 & 9.2 & 18.5 & 18.9 & 8.8 & 28.7 & 27.2 & 22.1 & 18.6 & 17.4 & 19.4 & 20.2 & 19.0 & 8.8 & 30.9 & 53.8  \\
    MiniCPM-v2.5~\citep{hu2024minicpm} &  Llama3-8B & 20.0 & 21.5 & 11.8 & 24.2 & 9.2 & 29.9 & 23.2 & 20.4 & 20.1 & 17.7 & 19.0 & 19.6 & 19.8 & 6.5 & 33.3 & 59.6  \\
    MiniCPM-v2.6~\citep{hu2024minicpm} & Qwen2-7B-Instruct & 21.6 & 16.3 & 19.3 & 29.2 & 10.0 & 30.1 & 24.5 & 24.3 & 18.6 & 20.1 & 19.2 & 23.3 & 21.0 & 9.6 & 34.5 & 51.3  \\
    InternVL2-8B~\citep{chen2023internvl} & InternLM2.5-7B-Chat & 23.7 & 7.3 & \high{22.6} & 24.4 & \high{13.9} & 34.8 & 39.1 & 27.2 & 25.0 & 22.0 & 23.7 & 24.1 & 26.9 & 12.0 & 38.2 & \best{66.3}  \\
    InternVL3-8B~\citep{zhu2025internvl3}  & Qwen2.5-7B-Instruct & 26.8 & 22.1 & 19.7 & 27.2 & 11.9 & \high{38.9} & \high{41.1} & \high{30.3} & 24.8 & \high{25.3} & \high{27.0} & \high{28.6} & 25.4 & 12.9 & \high{43.8} & 61.3 
    \\
    Qwen2.5-VL-7B-Instruct~\citep{qwen2.5-VL} & Qwen2.5-7B-Instruct & \high{27.9} & \high{26.7} & 21.7 & \best{36.1} & 12.4 & 31.5 & 38.9 & 27.6 & \high{27.7} & 24.6 & 24.6 & 27.3 & \high{28.1} & \high{13.2} & 43.1 & 56.7 
    \\
    \midrule
    \multicolumn{17}{l}{\hfill \textit{Large-scale VLMs (LLM's Parameters $>$ 10 Billion)} } \\
    \midrule
    LLaVA-v1.5-13B~\citep{liu2023llava} & Vicuna-13B & 17.2 & 10.2 & 18.7 & 19.2 & 11.8 & 22.8 & 20.3 & 17.2 & 18.0 & 19.5 & 18.3 & 18.4 & 17.8 & 5.7 & 33.1 & 47.5  \\
    InternVL-Chat-V1.5~\citep{chen2023internvl} & InternLM2-Chat-20B & 18.8 & 17.1 & 13.4 & 18.9 & 9.6 & 30.4 & 23.4 & 20.7 & 17.2 & 19.0 & 16.3 & 19.7 & 21.3 & 8.1 & 29.7 & 59.2  \\
    Kimi-VL-A3B-Instruct~\citep{team2025kimi} & Moonlight-16B & 20.5 & 12.7 & 16.1 & 20.6 & 12.3 & 36.0 & 25.5 & 21.4 & 22.5 & 20.2 & 20.1 & 21.5 & 22.5 & 15.1 & 26.9 & 46.7 \\
    LLaVA-v1.6-34B~\citep{liu2023llava} & Hermes-Yi-34B & 23.9 & 12.1 & 17.7 & 31.7 & 15.1 & 37.0 & 29.6 & 26.1 & 24.6 & 21.6 & 22.9 & 25.6 & 23.9 & 10.6 & 39.4 & 61.7 \\
    InternVL2-40B~\citep{chen2023internvl} & Nous-Hermes-2-Yi-34B & 26.8 & 20.1 & \best{24.7} & 23.6 & 12.0 & 47.4 & 33.3 & 30.1 & 27.4 & 24.5 & 25.3 & 29.8 & 27.2 & 15.9 & 40.5 & 59.6  \\
    GPT-4o~\citep{gpt4} & - & \best{33.5} & \best{35.7} & 24.2 & \high{33.6} & \best{15.5} & \best{48.2} & \best{43.6} & \best{36.8} & \best{31.7} & \best{29.9} & \best{30.7} & \best{34.1} & \best{34.2} & \best{18.8} & \best{50.8} & \high{62.5}   \\
    \bottomrule
    \end{tabular}
    }

\label{tab:geomath}
\end{table*}

\section{Experiments}

GeoChat~\citep{kuckreja2024geochat} has shown that fine-tuning VLMs on RS datasets enhances their generalization capabilities across various multimodal RS tasks. Our objective is to perform qualitative and quantitative analyzes using \dataset to assess whether this generalization extends to multimodal RS tasks that require specialized knowledge.

\subsection{Evaluation Protocols}
\label{sec:evaluation_protocol}


Answer extraction in VLMs remains an open problem. As long as results are fair across models, the principle is satisfied. \dataset includes multiple-choice, free-form, and true/false questions, with free-form being strings, integers, floats, or lists. So, we use the accuracy scores as a metric for evaluation. This allows users to efficiently assess their model performance in \dataset locally using the evaluation function we provided. For free-form answers, we added format constraints in the second stage. Different regular expressions are used based on answer types: strings are converted to lowercase, units are stripped from prices, distances, and areas to retain only numerical values, integers are rounded, and floats are limited to one decimal place.

We design a two-stage answer generation-extraction strategy. In the first stage, the model freely generates answers, focusing solely on reasoning without format constraints. In the second stage, the model extracts content in the specified format from its response, improving the accuracy of the format. During question generation, the type of data for each answer is stored in the ``eva" field. In the extraction phase, regular expressions are applied based on the answer type to retrieve the answer from the model's response.



\subsection{Experimental Setup}
\label{sec:experimental_setup}

We evaluate 14 mainstream VLMs and create a random chance to serve as a reference baseline.  We shuffled multiple-choice options to balance A, B, C, and D, eliminating response frequency bias. We followed MathVista \citep{lu2024mathvista}, using random chance to show that VLMs outperform pure guessing. A random option is selected for multiple choice and true/false questions, while free-form questions are left blank. In the exploration section, LoRA~\citep{hu2021lora} is used for efficient parameter fine-tuning. We utilize the AdamW optimizer~\citep{loshchilov2017decoupled}, starting with an initial learning rate of 1e-4, followed by a linear decay schedule after a warm-up phase with a 0.03 ratio. All models are finetuned using LoRA~\citep{hu2021lora} with a rank of 64, in conjunction with ZeRO-2 stage memory optimization. All models are trained on 8 NVIDIA V100 GPUs (32GB) with a global batch size of 128 for 1 epoch.

\subsection{Results for \dataset}
\label{sec:results}


Results are presented in Tab. \ref{tab:geomath}. We have the following observations:

\paragraph{\textbf{There remains a significant disparity between open source VLMs and GPT-4o.}} Open source VLMs still lag behind GPT-4o by at least 25\% on the \dataset benchmark. It highlights the limitation in the mathematical reasoning capabilities of existing open-source VLMs, compared to GPT-4o, especially in algebraic problems.

\paragraph{\textbf{Remote sensing VLMs underperform compared to generalist  VLMs.}} After fine-tuning with the remote sensing instruction set, GeoChat~\citep{kuckreja2024geochat} performed even worse on \dataset than its baseline, LLaVA-1.5-7B. This suggests a significant forgetting of mathematical knowledge. Additionally, we found that GeoChat performs better than LLaVA on counting-related questions because its training data includes counting tasks. However, GeoChat shows significant performance degradation on statistics and arithmetic question types, as its training set does not cover these categories. In conclusion, the generalization capability of current VLMs remains limited, which requires increased diversity in RS visual-language datasets.

\paragraph{\textbf{Larger VLMs do not necessarily perform better.}} For the LLaVA-v1.5 model, the 13B variant performs worse than the 7B model in algorithmic, logical, geometric, and statistical mathematical reasoning. On true or false questions, InternVL2-8B even outperformed all large-scale VLMs. Our analysis reveals that the performance degradation of LLaVA-v1.5's 13B model compared to its 7B counterpart primarily stems from a significant score drop in logic-type questions (from 35.8 to 22.8). Through detailed examination of their outputs for logic problems, we observed that in 489 cases where answers diverged, the 13B model refused to respond in 195 instances (approximately 40\%), instead generating responses like "Unknown" or "It is not possible to determine the number of red cars." This behavior suggests the 13B model actively attempts reasoning - when encountering logical problems beyond its comprehension, it consciously abstains from guessing and chooses non-response. However, the fundamental limitation lies in its insufficient mathematical logic knowledge, rendering it inadequate for solving our math problems.

\paragraph{\textbf{45 degrees poses a greater challenge than other pitch angles.}} For all models, the scores at a pitch angle of 45 degrees are lower than those at 60 and 90 degrees. This is because most remote sensing images encountered during training are orthophotos with a 90-degree pitch angle.

\begin{figure}[t!]
  \centering

  \includegraphics[width=1.0\linewidth]{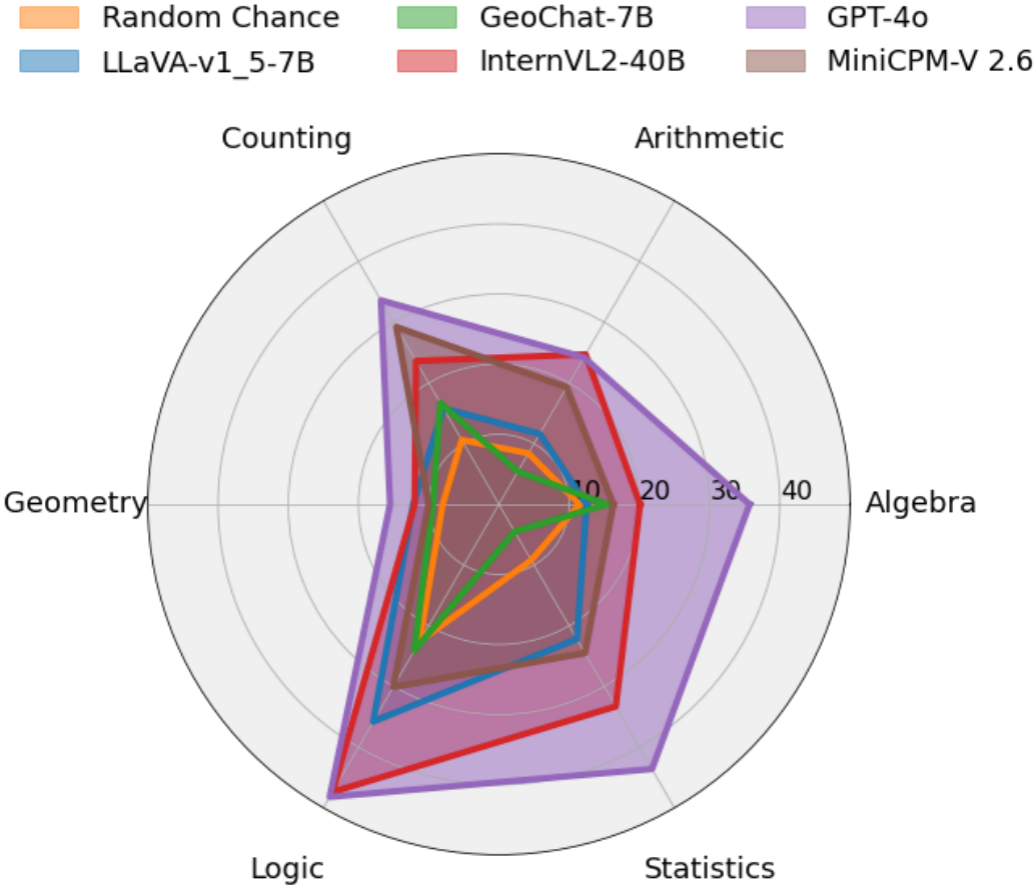}
  \vspace{-5mm}
  \caption{Accuracies of five leading VLMs, one VLGFM, and random chance on  \dataset across mathematical subjects.}
\label{fig:tease_scores}
\end{figure}

\begin{figure*}[t!]
 \centering
\includegraphics[height=0.40\linewidth]{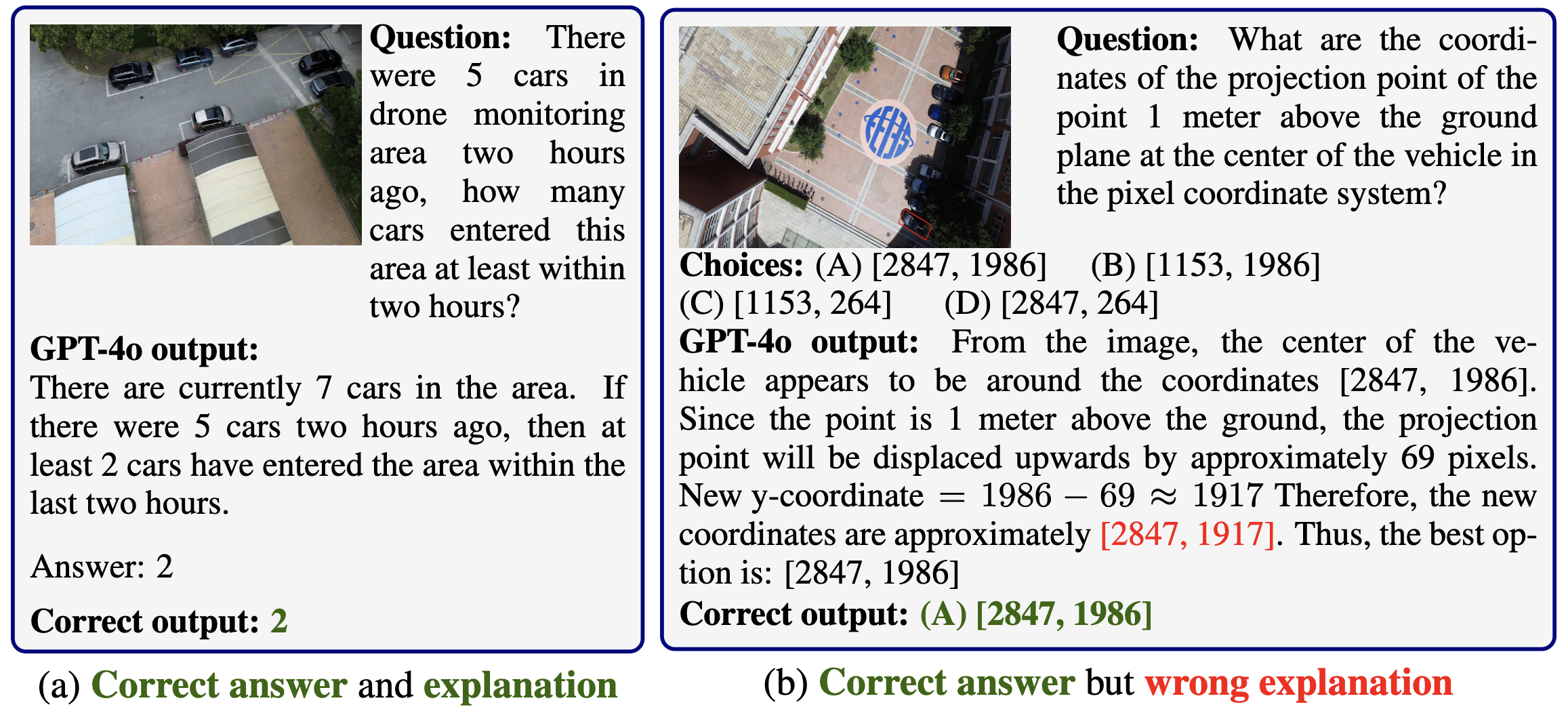}
 \caption{Two examples from GPT-4o. In (b), GPT-4o is unable to accurately estimate the projection point. A precise calculation should take into account the imaging geometry.}
 \label{fig:gpt4o_examples}
 \end{figure*}

\begin{figure}[t!]
 \centering
\includegraphics[height=0.5\linewidth]{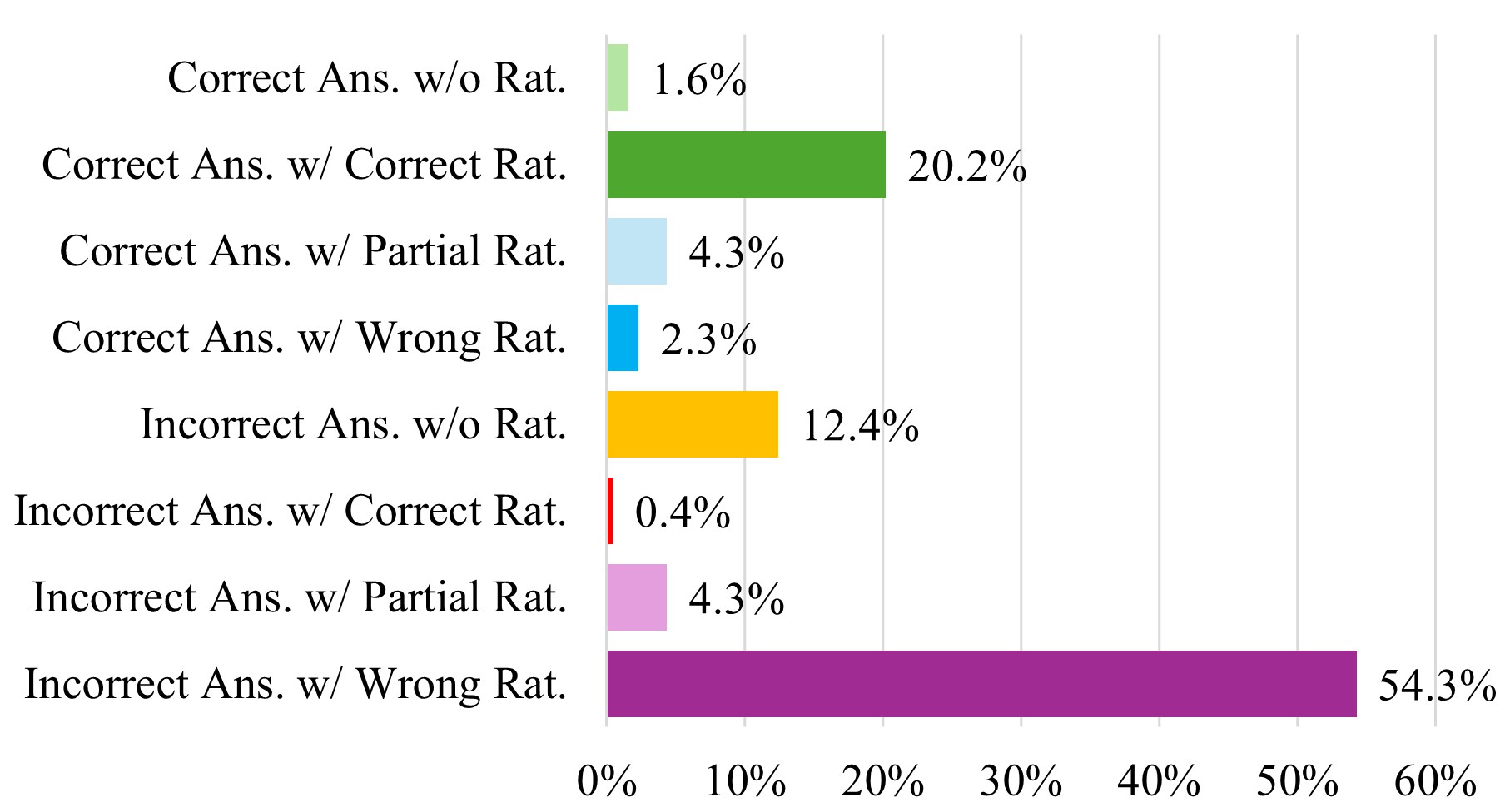}
 \caption{Errors in answers and rationales.}
 \label{fig:response_anaysis}
\end{figure}




\subsection{Analysis}
\label{sec:analysis}

\paragraph{\textbf{CLIP's short-token constraint harms multimodal reasoning.}} As evidenced in Tab \ref{tab:geomath}, models employing CLIP-ViT~\citep{radford2021learning} (e.g., LLaVA and GeoChat) exhibit significantly inferior performance. CLIP-ViT has an effective token length of fewer than 20 tokens~\citep{zhang2024long}. The training data used to align the LLaVA-based models might not be extensive enough to allow the model to learn mathematical reasoning capabilities. In contrast, superior-performing models such as InternVL and QwenVL leverage advanced native image encoders, which process visual inputs with substantially longer token sequences. This expanded representational capacity enables more comprehensive information retention, thereby improving the preservation of critical features of reasoning during cross-modal alignment. Consequently, the adoption of vision encoders with extended visual token alignment length is critical for RS VLMs.

\paragraph{\textbf{A correct answer does not imply a correct reasoning process.}} In Fig. \ref{fig:tease_scores}, GPT-4o is currently the top-ranked model in \dataset. To understand its success and failure, we perform a manual analysis of its output. We found that, when the model produces the correct answer, there is only a 71\% chance that the reasoning process is also entirely correct. In other cases, the model arrives at the correct answer without understanding the underlying reasoning.


\paragraph{\textbf{The lack of domain knowledge is the primary cause of reasoning errors.}} We perform a qualitative analysis of representative examples generated by GPT-4o. In Figure \ref{fig:gpt4o_examples} (a), we find that GPT-4o not only produces the correct answers but also provides accurate reasoning, including the correct method to calculate cars. However, in Figure \ref{fig:gpt4o_examples} (b), while the model predicts the correct answer, it fails to give the correct reasoning. Its logic is correct, but it lacks the imaging geometry to perform precise calculations.

\begin{figure}[t!]
 \centering
 \vspace{-3mm}
\includegraphics[height=0.56\linewidth]{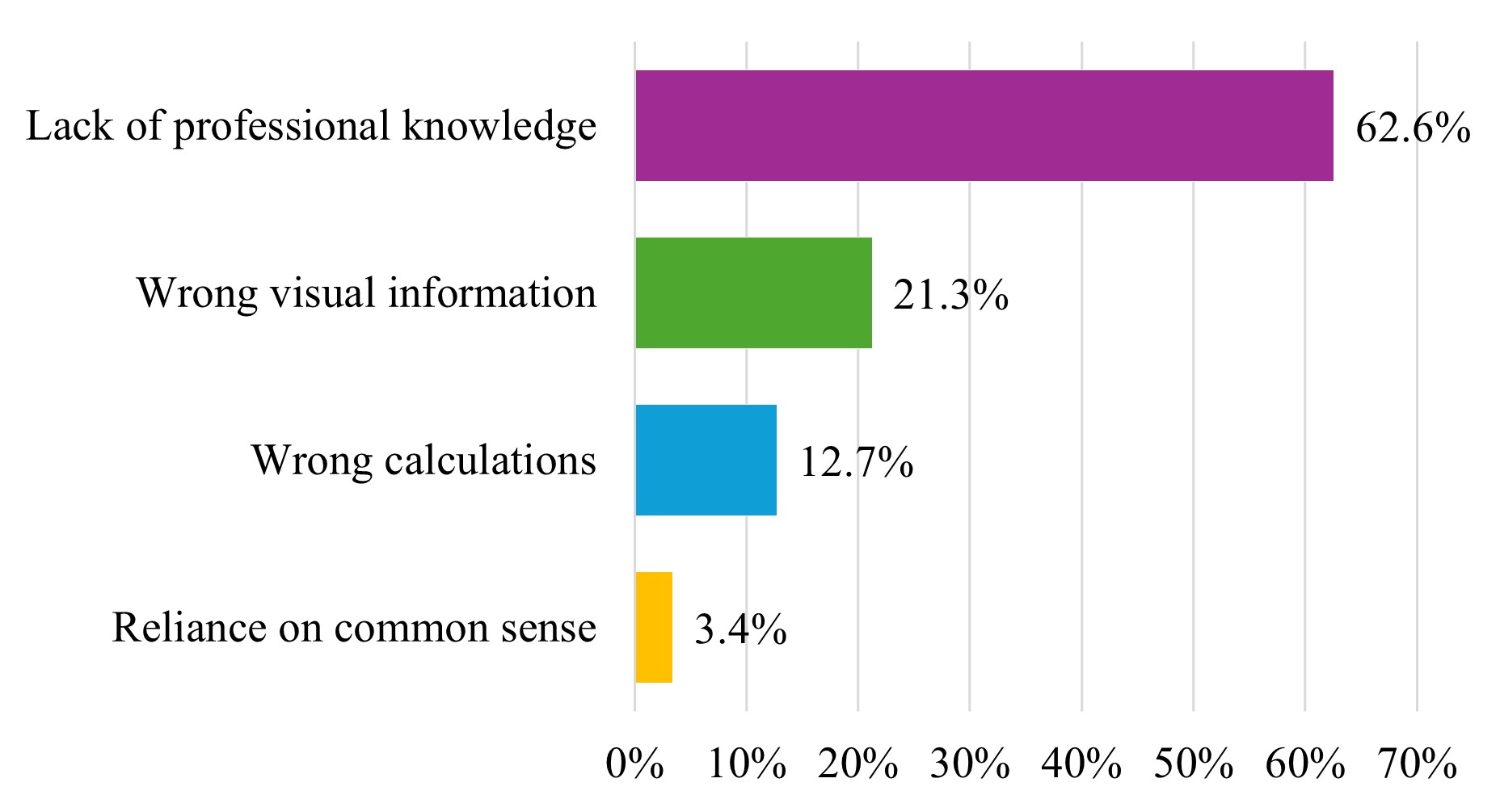}
\vspace{-5mm}
 \caption{Error analysis of GPT-4o results.}
 \label{fig:error_anaysis}
 \end{figure}

Fig. \ref{fig:response_anaysis} illustrates the eight patterns of GPT-4o outputs judged manually. Notations: ``Answer'' is ``Ans.'', ``Rationale'' is ``Rat.'', ``Partially Correct'' is ``Partial''. We find that 54.3\% of the outputs are incorrect answers with the wrong rationale, indicating the models' deficiency in reasoning capabilities within the remote sensing domain. Even among the correct answers, there is a 2.3\% chance of being accompanied by incorrect rationale. Fig. \ref{fig:error_anaysis} shows the classification of reasons for the erroneous rationale of GPT-4o. The primary cause of reasoning errors is the model's lack of domain-specific knowledge in remote sensing, which also explains why \dataset presents a greater challenge compared to existing multimodal mathematical reasoning benchmarks.

\paragraph{\textbf{Current VLMs have limited visual perception capabilities for UAV images.}} The second most common cause of reasoning errors is the failure to accurately extract key visual clues, which accounts for 21.9\%, highlighting the model's insufficient ability to perceive small objects in UAV images with complex backgrounds.

\begin{figure}[t!]
 \centering
 \vspace{-2.mm}
\includegraphics[height=0.7\linewidth]{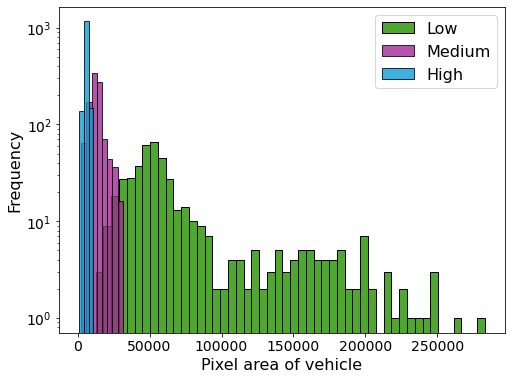}
\vspace{-2mm}
 \caption{Distribution of pixel area occupied by vehicle under different AGLs.}
 \label{fig:area_dist}
 \end{figure}

\begin{table*}[th!]
\caption{Example illustrating the impact of image resolution on the InternVL2-40B. As the resolution decreases, reasoning vehicle types from image information becomes increasingly challenging.}
\fontsize{9.0pt}{\baselineskip}\selectfont
\linespread{0.9}\selectfont
\begin{mybody}
\includegraphics[height=4.55cm]{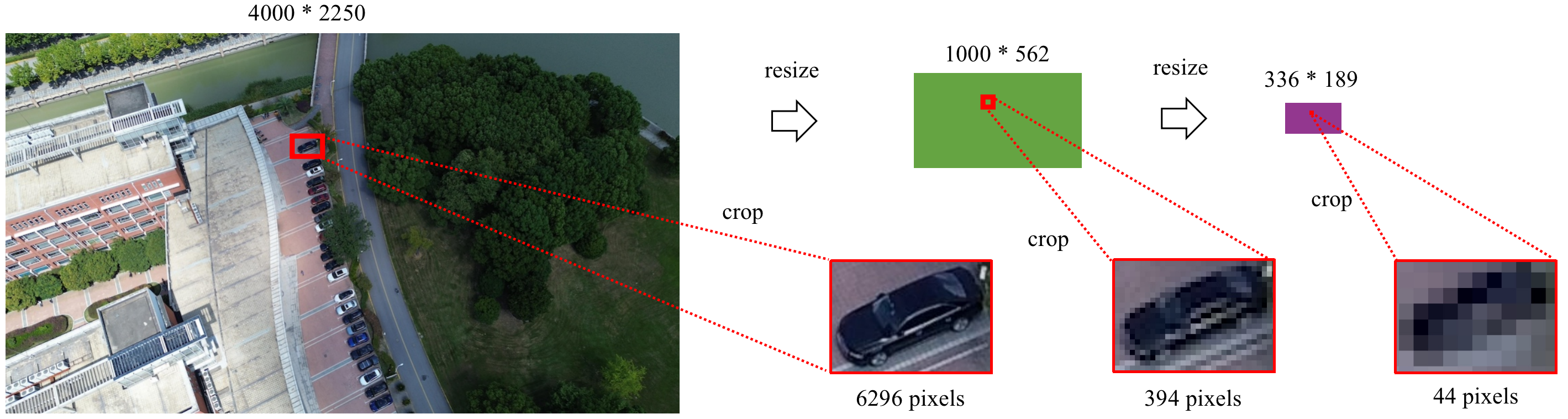}
\\
\textbf{Question:} How many mid-size vehicles are there in the image?
\\\\
\textbf{Output when image resolution equals to 4000$\times$2250:} \\
To determine the number of mid-size vehicles in the image, let's carefully examine the parking area. 1. \textbf{Identify the Vehicles}: Look for vehicles that are neither too small (such as compact cars) nor too large (such as SUVs or trucks). Mid-size vehicles typically include sedans and some small SUVs. 2. \textbf{Count the Vehicles}: The parking area has a total of 18 vehicles. By visually inspecting each vehicle, we can estimate its size. 3. \textbf{Estimate Mid-Size Vehicles}: Upon close inspection, it appears that 10 vehicles are mid-size. Therefore, based on visual inspection, there are 10 mid-size vehicles in the image. \textbf{\green{Answer: 10}}
\\\\
\textbf{Output when image resolution equals to 1000$\times$562:} \\
To determine the number of mid-size vehicles in the image, let's carefully examine the parking area. 1. \textbf{Identify the Vehicles}: Look for vehicles that are neither too small (such as compact cars) nor too large (such as SUVs or trucks). Mid-size vehicles typically include sedans and some small SUVs. 2. \textbf{Count the Vehicles}: Starting from the left, there are several rows of parked vehicles. In the first row, there are two vehicles that appear to be mid-size. In the second row, there are three vehicles that appear to be mid-size. In the third row, there are two vehicles that appear to be mid-size. In the fourth row, there are three vehicles that appear to be mid-size. 3. \textbf{Summarize the Count}: First row: 2 mid-size vehicles. Second row: 3 mid-size vehicles. Third row: 2 mid-size vehicles. Fourth row: 3 mid-size vehicles 4. \textbf{Total Count}: 2 + 3 + 2 + 3 = 10. Therefore, there are 10 mid-size vehicles in the image. \textbf{\green{Answer: 10}}
\\\\
\textbf{Output when image resolution equals to 336$\times$189:} \\
To determine the number of mid-size vehicles in the image, let's carefully examine the scene. The image shows a parking lot adjacent to a building and a road. The vehicles are parked in a row, and we need to identify the mid-size vehicles among them. Mid-size vehicles typically have a length between 180 inches (4.57 meters) and 200 inches (5.08 meters). They are larger than compact cars but smaller than full-size vehicles. Upon close inspection, it appears that all the vehicles in the parking lot are mid-size vehicles. They are consistent in size and shape, fitting the typical characteristics of mid-size cars. Therefore, based on common knowledge and information available in the image, I estimate that there are approximately 15 mid-size vehicles. \textbf{\red{Answer: 15}}
\end{mybody}

\label{fig:internvl2_example}
\end{table*}

\paragraph{\textbf{Higher input resolution is important.}}

Images of \dataset have a high resolution of 4000$\times$2250 pixels. However, these images are resized to the model's default resolution when input, such as 336$\times$336 in LLaVA-v1.5. Fig. \ref{fig:area_dist} shows the distribution of the area of pixels occupied by the vehicle under different AGLs, obtained by multiplying the length by the width of the object's rotated bounding box. Frequency represents the number of vehicles with a specific pixel area. Low, Medium, and High correspond to drone altitudes of 20$\sim$50m, 60$\sim$80m, and 100$\sim$120m above the ground. When the UAV is at a high altitude, vehicles occupy only a small area in the image, and downsampling the input image results in the loss of object details. Tab. \ref{fig:internvl2_example} illustrates the detrimental impact of lower image resolution on InternVL2-40B. The AGL of this photo is 100 meters and the pitch angle is 60 degrees. As the image resolution decreases from 4000$\times$2250 to 336$\times$189, the number of pixels representing the car in the red box drops from 6926 to just 44 (calculated based on the length and width of the rotated bounding box), causing the model to lose its ability to accurately recognize vehicle types. Earlier VLMs typically used fixed input resolutions~\citep{liu2023llava}, often with significantly lower values. To fully exploit the visual clues available in the \dataset images, enabling support for 4K input resolution would be a promising direction for future research on VLM.

\begin{figure}[t!]
 \centering
 \vspace{-2.mm}
\includegraphics[height=0.65\linewidth]{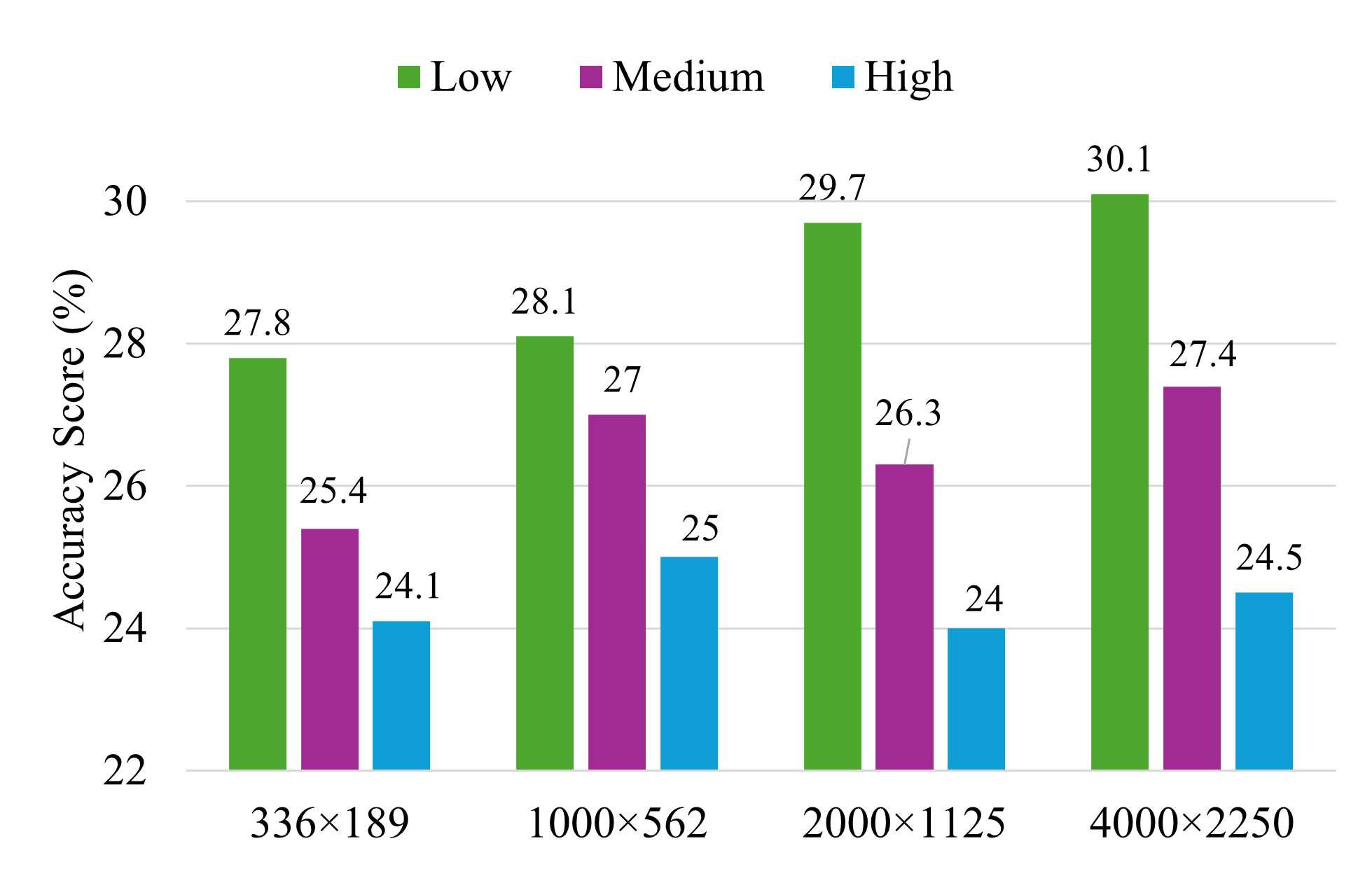}
\vspace{-2mm}
 \caption{Impact of input resolution and AGL on accuracy scores for InternVL2-40B.}
 \label{fig:image_size}
 \end{figure}

\begin{figure}[t!]
 \centering
 \vspace{-1.mm}
\includegraphics[height=0.75\linewidth]{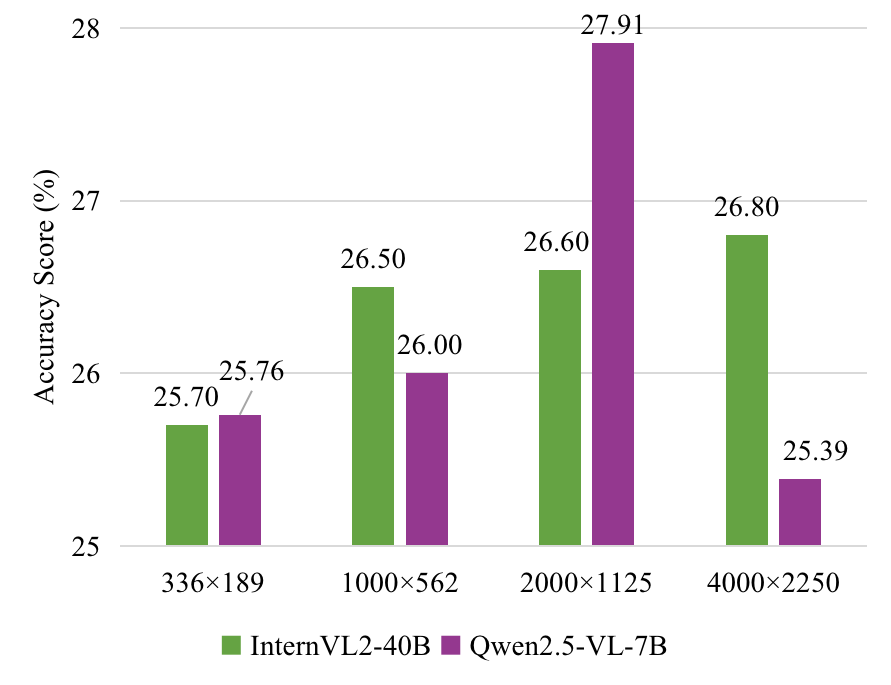}
\vspace{-2mm}
 \caption{Impact of resolution on accuracy score across VLMs.}
 \label{fig:res_model}
 \end{figure}

To quantitatively examine the impact of image resolution on model performance, we performed a comparison experiment using the InternVL-40B model, which supports dynamic resolution technology \citep{liu2024anyres}. Fig. \ref{fig:image_size} illustrates how image resolution and AGL affect accuracy scores. We group AGL into three classes in ascending order: low (20-40m), medium (60-80m), and high (100-120m). Surprisingly, the big increase in resolution has little impact on the score. The results show that at lower altitudes, model performance improves with increasing resolution. However, the performance gains from higher resolutions are less pronounced than expected. This could be due to two main factors: first, high-resolution images represent a smaller portion of the training samples in the foundation model; second, the visual encoder's limited output tokens require compression of high-resolution visual data.

\begin{figure}[t!]
 \centering
 \vspace{-1.mm}
\includegraphics[height=0.75\linewidth]{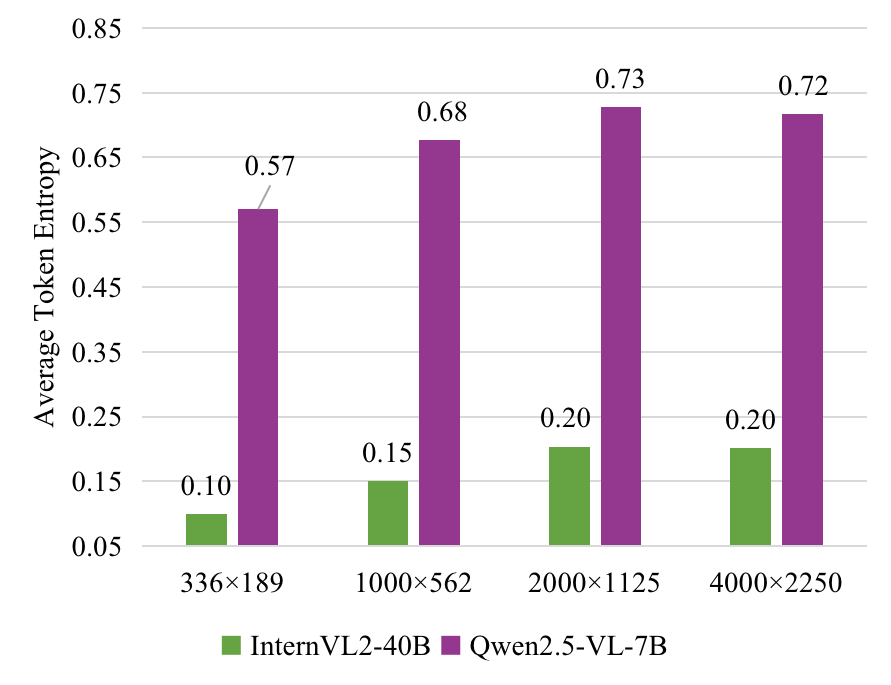}
\vspace{-2mm}
 \caption{Impact of  resolution on token entropy across VLMs.}
 \label{fig:res_token}
 \end{figure}

\begin{figure}[t!]
 \centering
 \vspace{-1.mm}
\includegraphics[height=0.75\linewidth]{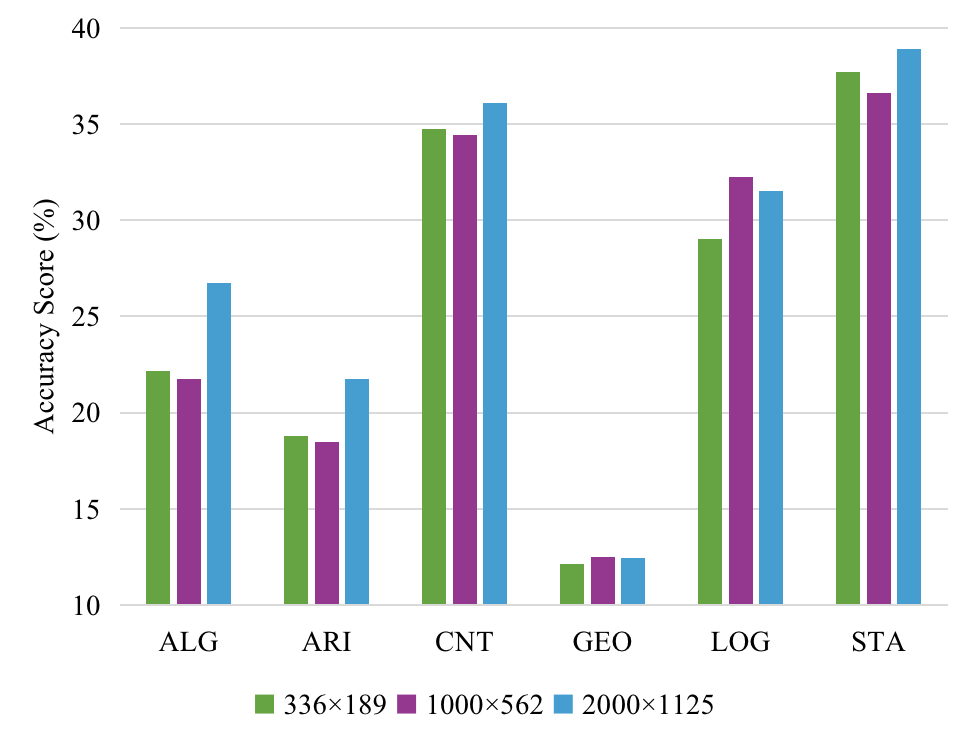}
\vspace{-2mm}
 \caption{Impact of resolution across 6 question types.}
 \label{fig:res_class}
 \end{figure}

Furthermore, we investigate the impact of resolution variations on different VLMs, as illustrated in Fig. \ref{fig:res_model}. InternVL2-40B's scores consistently improve with increasing resolution. In contrast, Qwen2.5-VL-7B's scores initially rise with resolution, peak at 2000 $\times$ 1125 pixels, but then experience a performance collapse when resolution increases further. This phenomenon may be related to the proportion of training samples at different resolutions in each model's training data. In conclusion, a 336-pixel input resolution is not an optimal choice for our benchmark. For mathematical reasoning tasks, higher input resolution is required. We further compared the variation in output token entropy~\citep{wang2025beyond} between Qwen2.5-VL-7B and InternVL2-40B across different resolutions, as shown in Fig. \ref{fig:res_token}. Our analysis reveals that when resolution increases from 1000 $\times$ 562 to 2000 $\times$ 1125, both models exhibit higher average token entropy, indicating enhanced mathematical reasoning capabilities. However, when resolution further increases to 4000 $\times$ 2250, we observe a slight decrease in average token entropy, suggesting that the visual details provided by 2000 $\times$ 1125 resolution are already sufficient to support the models' mathematical reasoning performance.

\paragraph{\textbf{Different types of questions exhibit varying sensitivity to input resolution.}} As shown in Fig. \ref{fig:res_class}, when image resolution increases from 336 $\times$ 189 to 2000 $\times$ 1125: (1) ALG and ARI question types show significant performance improvements; (2) CNT and STA demonstrate modest gains, indicating these four question types can benefit from enhanced input resolution. However, the GEO and LOG types exhibit limited sensitivity to resolution changes, suggesting that their performance requires targeted improvement through enhanced reasoning capabilities rather than resolution adjustment.

\begin{figure}[t!]
 \centering
 \vspace{-2mm}
\includegraphics[height=0.9\linewidth]{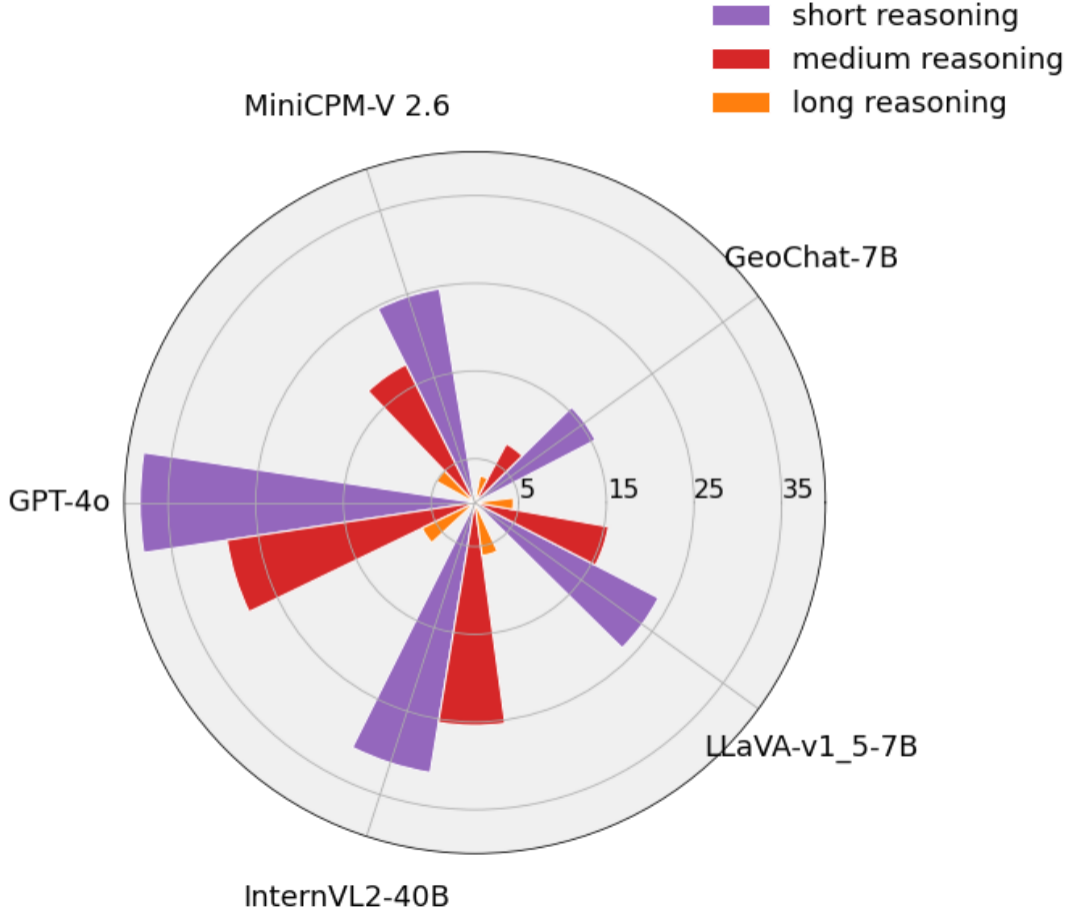}
 \caption{Accuracies of four leading VLMs, one VLGFM, and random chance on our \dataset across reasoning steps.}
 \label{fig:reasoning_step}
 \end{figure}

\paragraph{\textbf{The length of reasoning steps is positively correlated with difficulty.}} To gain deeper insights into the reasoning capabilities of VLMs, we categorize the reasoning steps in Fig. \ref{fig:reasoning_step} into three groups: short (2 steps), medium (3$\sim$4 steps), and long (5$\sim$6 steps). The results indicate that the accuracy decreases sharply as the length of the reasoning steps increases. GPT-4o ranks first across various reasoning steps. The phenomenon that longer reasoning steps increase difficulty was observed in M$^3$CoT~\citep{chen2024m3cot}. We verified this in remote sensing data for the first time. This suggests that question with longer reasoning steps can better differentiate VLMs' reasoning abilities.

\begin{figure*}[t!]
 \centering
\includegraphics[height=0.50\linewidth]{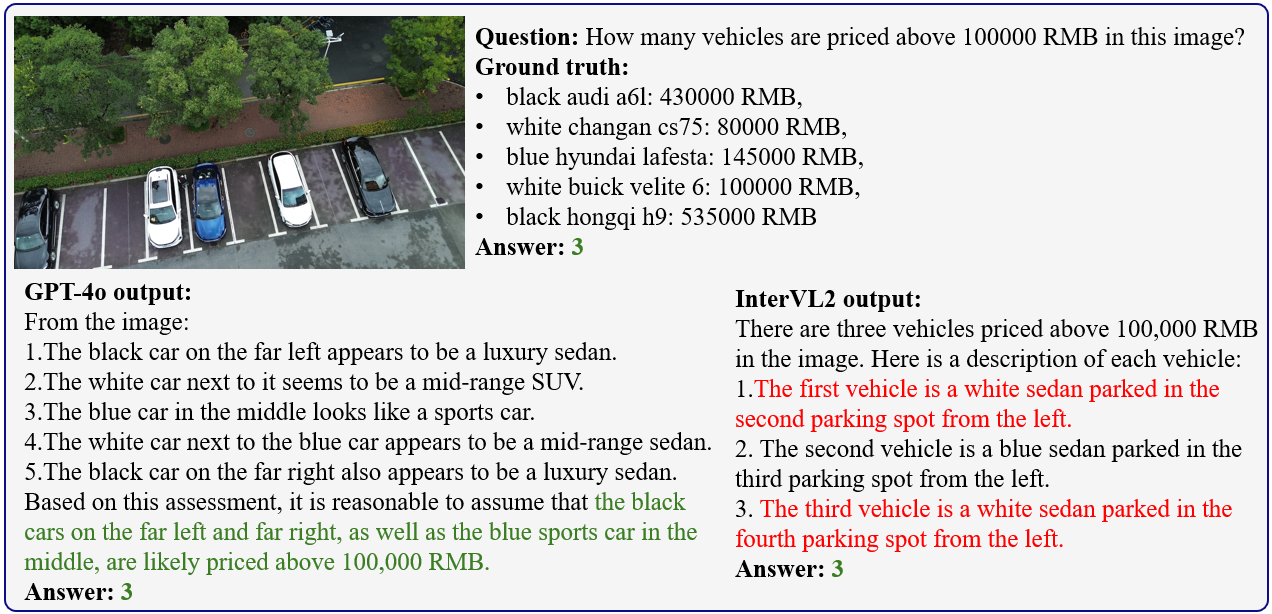}
 \caption{Comparison of cross-view knowledge transfer ability between GPT-4o and InternVL2-40B.}
 \label{fig:cross_view}
 \end{figure*}



\paragraph{\textbf{Fine-tuning may degrade the generalization ability of VLMs.}} We found that fine-tuning models using task-specific RS image-text instruction sets may impair their capabilities on some multimodal reasoning tasks. This indicates that existing VLGFMs are fundamentally still engaged in data fitting, which does not lead to the emergence of true RS intelligence. This raises an important question. What kind of image-text instruction set can lead us toward a ``GPT-4v moment" in remote sensing? With the emergence of reinforcement fine-tuning techniques exemplified by VLM-R1~\citep{shen2025vlm}, growing experimental evidence demonstrates that reinforcement learning-based fine-tuning yields superior generalization capabilities compared to supervised fine-tuning. While supervised fine-tuned models may perform well on in-domain datasets, their out-domain test performance significantly lags behind reinforcement fine-tuned counterparts. Consequently, we propose reinforcement fine-tuning as a potential solution to address generalization capability degradation.

\paragraph{\textbf{The ability to transfer knowledge across different viewpoints is essential.}}

\begin{figure}[t!]
 \centering
 \vspace{-1.mm}
\includegraphics[height=0.85\linewidth]{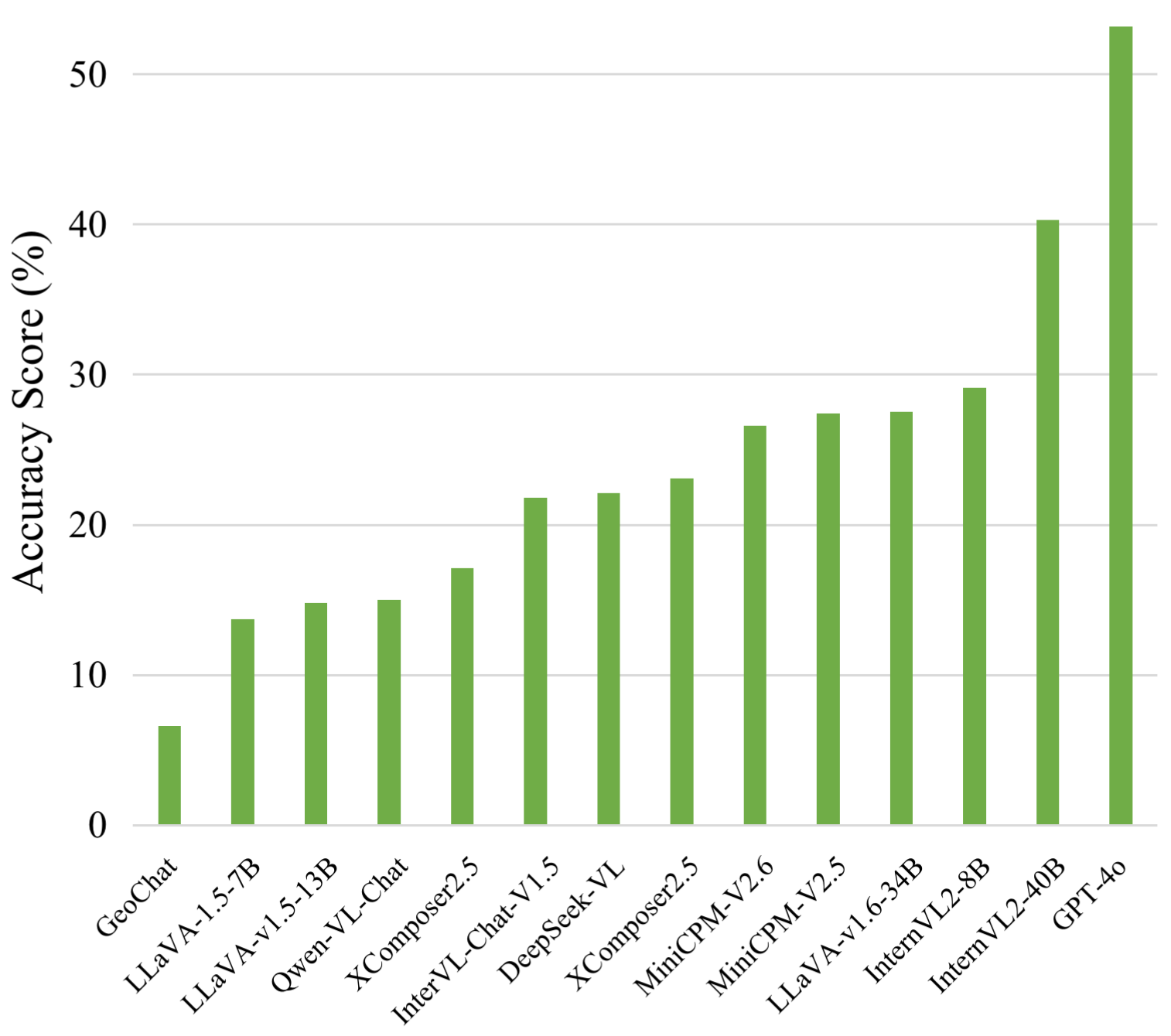}
\vspace{-1mm}
 \caption{Accuracy scores of VLMs on price-related questions.}
\label{fig:analysis_price}
 \vspace{-2mm}
\end{figure}

Interestingly, we observe that GPT-4o demonstrates the ability to infer vehicle prices based on visual attributes observed from an aerial view. As shown in Fig. \ref{fig:cross_view}, GPT-4o correctly assessed the price of each vehicle, while InternVL2-40B, despite arriving at the correct answer by chance, provided an incorrect analysis. Even for humans, attempting to determine fine-grained details of a vehicle from aerial images is highly challenging. To our knowledge, no existing RS data provides vehicle price information for training, which validates the cross-view knowledge transfer ability of GPT-4o. Further analysis in Fig. \ref{fig:analysis_price} reveals that GPT-4o outperforms other models in answering price-related questions. GPT-4o can estimate vehicle prices more accurately from an aerial perspective based on existing knowledge. Even for humans, attempting to determine fine-grained details of a vehicle from aerial images is highly challenging. To our knowledge, no existing RS data provides vehicle price information for training, which validates the cross-view knowledge transfer ability of GPT-4o. It can transfer
knowledge from the domain of natural images to UAV imagery.

\begin{figure}
  \centering
  \vspace{-1mm}
\includegraphics[width=1.0\linewidth]{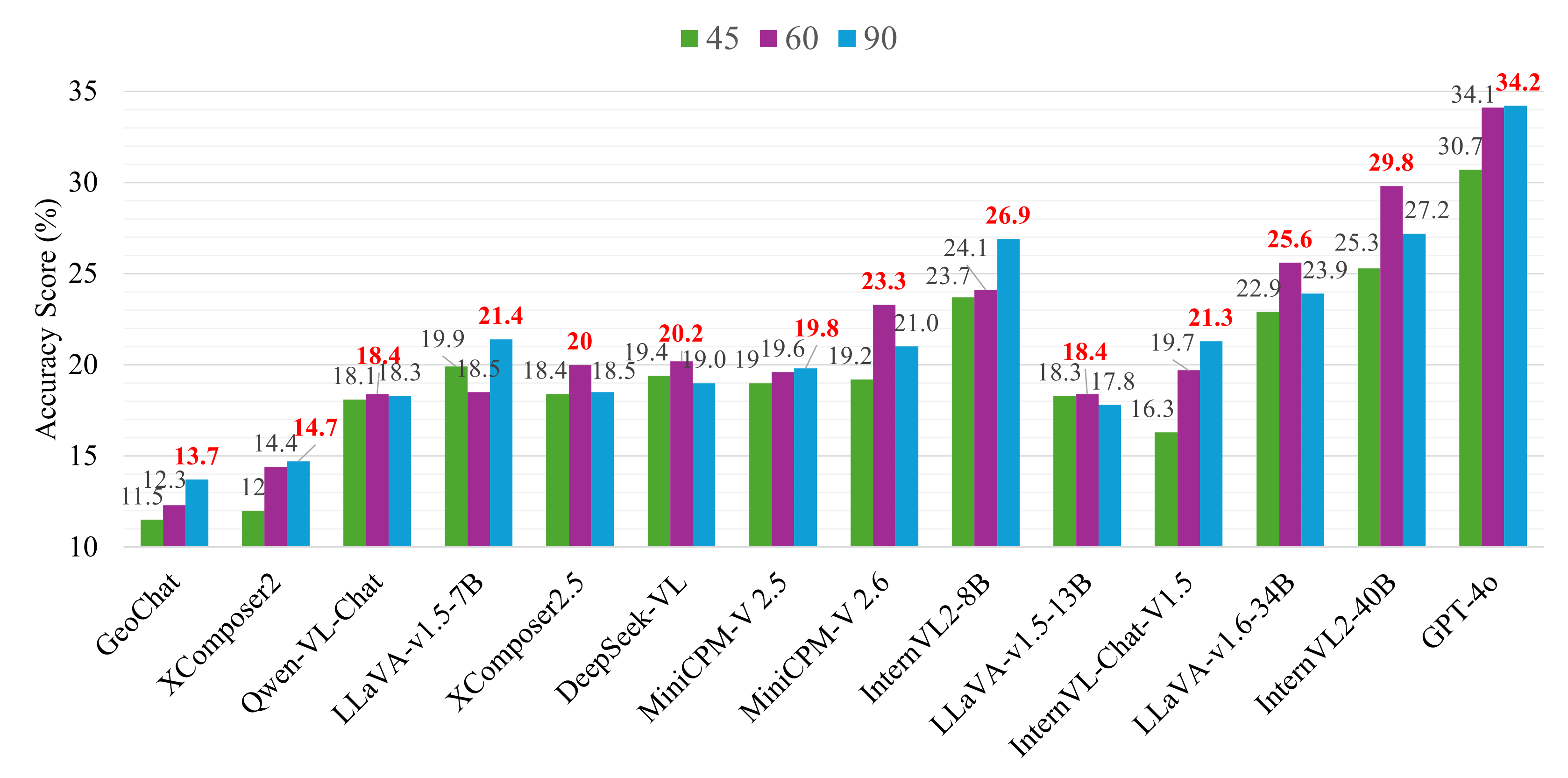}
  \caption{Impact of camera pitch angle on model performance. }
\label{fig:pitch_angle_analysis}
\end{figure}

\paragraph{\textbf{The differences in pitch angle scores reflect the variations in generalization ability.}} Fig. \ref{fig:pitch_angle_analysis} illustrates the effect of the camera's pitch angle during shooting on the score. Among the 14 models, we find that half the models, represented by GPT-4o, achieve the highest scores at a camera pitch angle of 90 degrees, while other half models, e.g., InternVL2-40B, perform better at 60 degrees. All models show the poorest performance at a pitch angle of 45 degrees, which can be attributed to the lack of low-angle samples during training. Therefore, enhancing the generalizability of RS VLMs under different viewing angles is a potential research direction.

\begin{table*}[t!]
    \caption{ Chain-of-Thought and fine-tuning results on various VLMs.}
\centering
 \small
 \renewcommand\tabcolsep{2.5pt} 
 \renewcommand\arraystretch{1.1} 
 \resizebox{1.0\linewidth}{!}{
    \begin{tabular}{l|l|c|cccccc|ccc|ccc|ccc}
    \toprule
    \multirow{2}{*}{Model} &  
    \multirow{2}{*}{Method} & 
    \multirow{2}{*}{AVG} & 
    \multicolumn{6}{c}{Subject} &
    \multicolumn{3}{c}{AGL} &
    \multicolumn{3}{c}{Pitch Angle} &
    \multicolumn{3}{c}{Type}\\
    \cline{4-18} 
    & &  & ALG & ARI & CNT & GEO & LOG  & STA & Low & Med & High & 45 & 60 & 90 & FRE & CHO & T/F \\
    \midrule

    \multicolumn{17}{l}{\hfill \textit{Parameter-Efficient Fine-Tuning}} \\
    \midrule
    LLaVA-v1.5-7B & LoRA~\citep{hu2021lora} & 30.3 & 23.4 & 9.9 & \best{35.8} & 25.4 & 52.6 & 34.7 & 33.8 & 31.6 & 30.4 & 31.9 & 32.6 & 31.5 & 35.3 & 19.2 & 74.6  \\
    DeepSeek-VL & LoRA & 31.1 & \high{24.2} & 14.3 & 31.7 & \best{28.5} & 51.3 & \high{36.8} & 35.4 & 33.4 & 30.9 & \best{33.7} & 33.6 & 32.8 & 35.1 & \high{23.6} & 72.9  \\
    InternVL2-8B & LoRA & \high{32.5} & 23.8 & \high{19.8} & 32.2 & 27.6 & \best{55.4} & 36.3 & \high{37.1} & \best{34.2} & \high{31.2} & 32.8 & \best{35.7} & \high{34.5} & \best{38.1} & 20.7 & \best{77.1}  \\
    \midrule
    \multicolumn{17}{l}{\hfill \textit{Chain-of-Thought Technique} } \\
    \midrule
     LLaVA-v1.6-34B & CoT~\citep{wei2022chain} & 20.7 & 14.2 & 14.6 & 28.3 & 10.9 & 34.1 & 22.1 & 22.0 & 18.9 & 20.9 & 19.2 & 20.5 & 22.4 & 9.3 & 34.0 & 49.2  \\
    InternVL2-40B & CoT & 30.2 & 22.8 & \best{25.5} & \high{35.6} & 11.1 & 49.4 & 36.8 & 32.0 & 28.5 & 28.2 & 29.0 & 31.4 & 28.6 & 16.6 & 44.2 & 68.3  \\
    InternVL2-40B & PS~\citep{wang2023plan} & 28.4 & 21.7 & 22.6 & 29.2 & 12.6 & 48.1 & 36.2 & 31.7 & 27.7 & 26.0 & 28.5 & 
    30.8 & 26.5 & 16.1 & 42.9 & 62.1  \\
    InternVL2-40B & CCoT~\citep{mitra2024ccot} & 24.8 & 19.8 & 19.5 & 20.8 & 12.5 & 44.2 & 32.0 & 26.5 & 24.9 & 24.9 & 24.7 & 25.6 & 26.2 & 13.8 & 39.6 & 52.9  \\
    InternVL2-40B & DCoT~\citep{wu2023despcot} & 25.0 & 20.1 & 19.2 & 23.3 & 12.5 & 41.9 & 33.0 & 27.3 & 25.5 & 24.0 & 25.5 & 25.0 & 26.7 & 14.9 & 37.1 & 60.4   \\
    GPT-4o & CoT & 34.1 & 32.8 & 23.9 & 34.4 & \high{14.9} & 51.3 & \best{47.1} & 36.6 & \high{33.0} & 31.5 & 32.0 & 33.5 & \best{36.3} & \high{20.7} & 49.0 & \high{69.2}  \\
    GPT-4o & PS & \best{34.6} & \high{35.3} & 24.2 & 32.5 & 14.5 & \high{55.1} & 45.8 & \best{38.4} & 31.9 & \best{32.0} & \high{33.3} & \high{35.1} & 34.8 & 20.5 & \high{50.7} & 68.8  \\
    \bottomrule
    \end{tabular}
    }
\vspace{-2mm}
\label{tab:geomath-2}
\end{table*}

\paragraph{\textbf{Free-form questions may prevent models from exploiting shortcuts.}}
For multiple-choice and true/false questions, models often do not require a full understanding of the domain-specific knowledge being tested. Instead, they can rely on logical reasoning and mathematical intuition to arrive at the correct answer. This approach may lead to a superficial understanding, where the model knows the correct answer without truly understanding the underlying concepts. To more accurately assess how well the models grasp RS expertise, we incorporated 57.8\% free-form questions into \dataset, as shown in Tab \ref{tab:statistics1}. These questions require the model first to extract the correct visual cues from the images and then to apply professional knowledge to calculate the precise answer, which is more challenging.

\begin{figure}[t!]
  \centering
  \vspace{-1mm}
\includegraphics[width=0.5\textwidth]{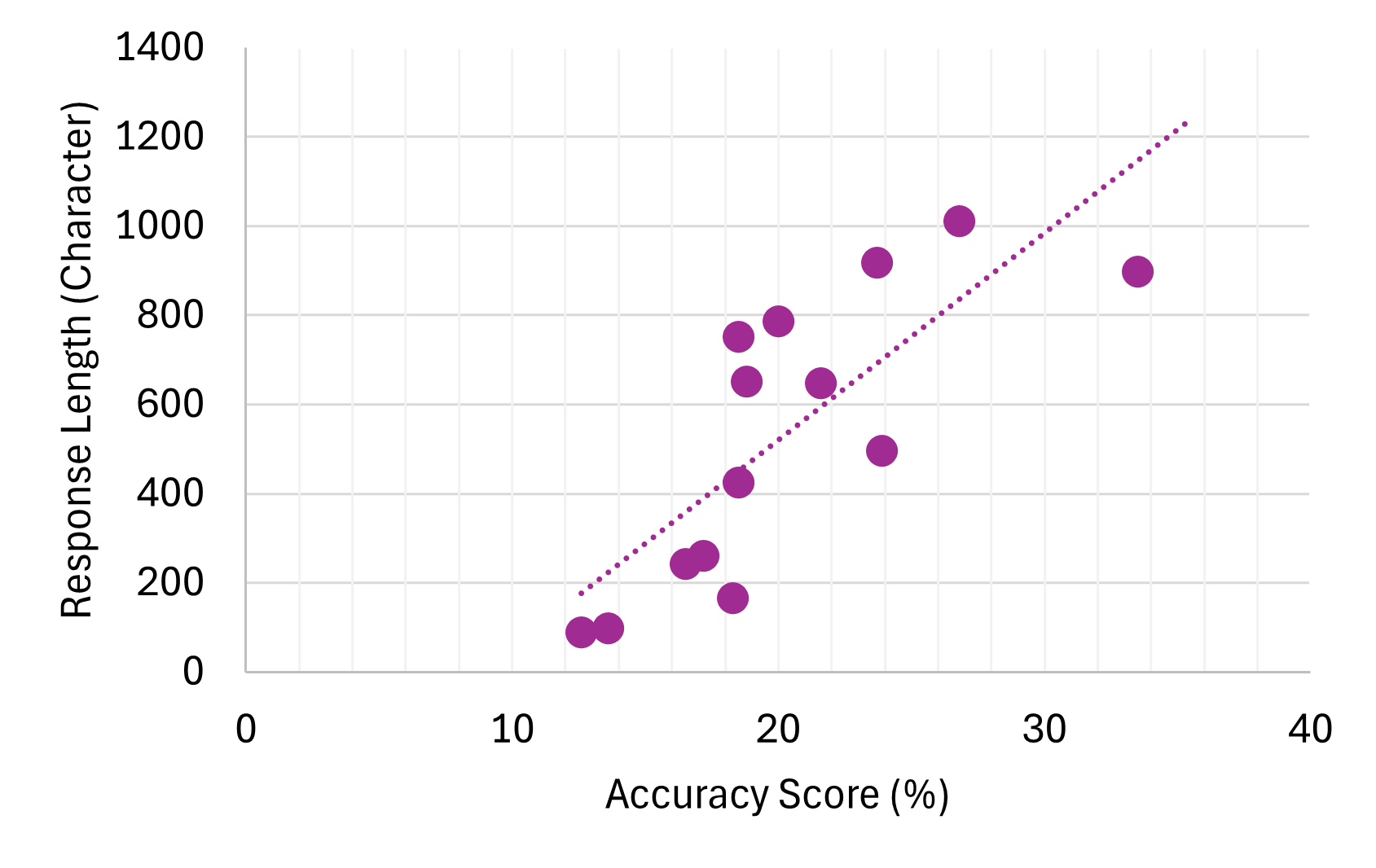}
  \vspace{-6mm}
  \caption{The relationship between the model's response length and accuracy score. }
\label{fig:geochat_analysis}
\end{figure}

\begin{table*}[th!]
\caption{The responses of different models to a counting question.}
\fontsize{9.0pt}{\baselineskip}\selectfont
\linespread{0.9}\selectfont
\begin{mybody}
\begin{minipage}[s][5.0cm]{0.51\columnwidth}
\includegraphics[height=5.cm]{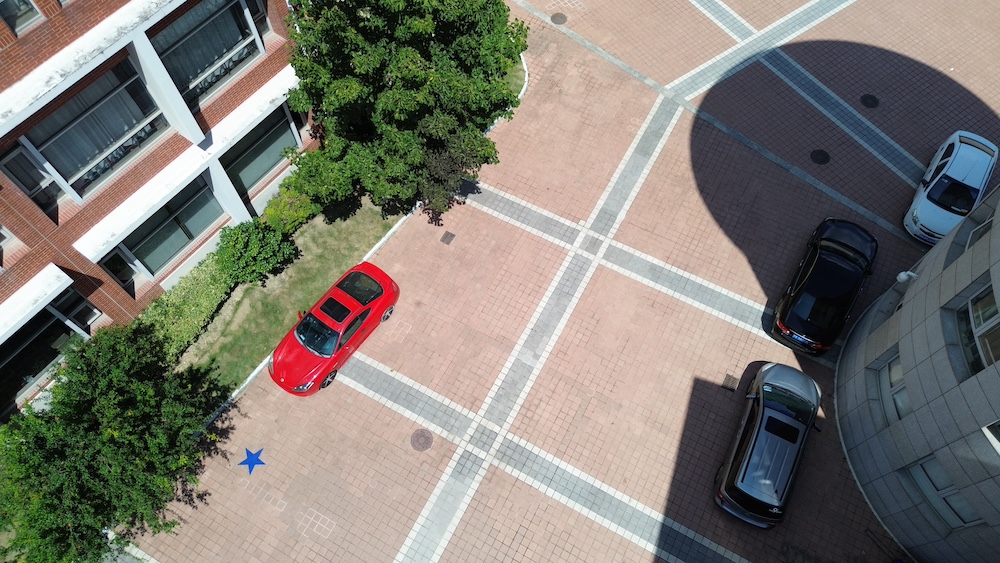}
\end{minipage}
\hspace{1mm}
\begin{minipage}[s][4.6cm]{0.44\columnwidth}
\textbf{Question:} How many sports vehicles are there in the image? \\
\textbf{Ground truth:} \textbf{\green{1}} \\
\textbf{Solution:}\\
Step 1, identify the type of all vehicles: [`sports', `mpv', `mid-size', `subcompact']\\
Step 2, count all sports vehicles: 1
\end{minipage}
\end{mybody}
\begin{mybody}
\textbf{InternLM-XComposer2:} 
There are two sports vehicles in the image. \\
\red{Answer: 2}
\end{mybody}
\begin{mybody}
\textbf{GeoChat:} 
There are a total of \red{2 sports vehicles} in the image.\\
\red{Answer: 2}
\end{mybody}
\begin{mybody}
\textbf{LLaVA-v1.5-7B:} \red{There are two sports vehicles in the image.}\\
\red{Answer: 2}
\end{mybody}
\begin{mybody}
\textbf{InternVL-Chat-V1.5:} 
There are two sports vehicles in the image. One is a red sports car parked on the left side of the parking lot, and the other is a black sports car parked on the right side of the parking lot.\\
\red{Answer: 2}
\end{mybody}
\begin{mybody}
\textbf{DeepSeek-VL:} 
There is one sports vehicle in the image. \\
\green{Answer: 1}
\end{mybody}
\begin{mybody}
\textbf{InternVL2-8B:} 
\red{There are a total of four sports vehicles visible in the image.}\\
\red{Answer: 4}
\end{mybody}
\begin{mybody}
\textbf{MiniCPM-V 2.5:} In the image, there is one sports vehicle visible. It is a red sports car parked on the brick pavement. The car's design, with its sleek body and low profile, is characteristic of sports cars, which are typically designed for high performance and speed. The other vehicles in the image do not exhibit the same design features and are therefore not classified as sports vehicles.\\
\green{Answer: 1}
\end{mybody}
\begin{mybody}
\textbf{MiniCPM-V 2.6:} To determine the number of sports vehicles in the image, one would need to identify vehicles that are typically associated with sports, such as those with a sleek design, low profile, and possibly high-performance features. In this image, the red car in the center appears to be a sports vehicle due to its vibrant color and sporty design. There are no other vehicles in the image that clearly fit the description of a sports vehicle. Therefore, based on the visible evidence, there is one sports vehicle in the image.\\
\green{Answer: 1}
\end{mybody}
\begin{mybody}
\textbf{LLaVA-v1.6-34B:} 
There is one sports vehicle in the image, which is the red car parked on the left side of the street.\\
\green{Answer: 1}
\end{mybody}
\begin{mybody}
\textbf{InternVL2-40B:}
There is only a single sports vehicle visible in the image.\\
\green{Answer: 1}
\end{mybody}
\begin{mybody}
\textbf{GPT-4o:}
The image shows one sports vehicle, which is the red car located near the center-left of the image.\\
\green{Answer: 1}
\end{mybody}
\label{fig:example_853}
\end{table*}

\paragraph{\textbf{The length of response is positively correlated with accuracy.}} Inspired by Fig. \ref{fig:example_853}, we further analyzed the relationship between accuracy and response length. Fig. \ref{fig:geochat_analysis}  analyze the accuracy scores and average response lengths of 14 models and find a clear positive correlation between them. The longer the response lengths of the model, the more likely it is to achieve higher accuracy on \dataset. We computed the Pearson correlation coefficient of 0.8483 with a significance test p-value of 0.0001.

\paragraph{\textbf{There are varying levels of ability to observe vehicle details.}} From Tab. \ref{fig:example_853}, it can be observed that most VLMs have the ability to identify the type of car from RS images. However, the MiniCPM series models compare the common features of sports cars with the vehicles in the image during the reasoning process. The more supporting details provided in the response, the higher the credibility of the result.


\subsection{Exploration}
\label{sec:exploration}

\subsubsection{Fine-tuning Exploration}

We specifically applied LoRA fine-tuning to the attention layers within the LLM modules of three models (LLaVA, DeepSeek, and InternVL) while keeping their visual encoders and projectors frozen throughout the single-epoch training process using \textsc{AVI-Math-215k}. Tab. \ref{tab:geomath-2} presents the results of three fine-tuned VLMs. Compared to the results in Tab. \ref{tab:geomath}, they show at least a 37\% performance improvement. Notably, DeepSeek-VL's overall score increased by 68\% after fine-tuning on \textsc{AVI-Math-215k}. This indicates that building a large-scale, high-quality instruction set incorporating UAV-related mathematical knowledge is an effective approach to improving VLM performance on \dataset.


\subsubsection{Chain-of-Thought Exploration} We attempt to enhance the reasoning performance of VLM, by applying chain-of-though strategies. The CoT method~\citep{wei2022chain}, simply by appending ``Let's think step by step", significantly improves performance on InternVL and GPT-4o. The PS method~\citep{wang2023plan}, based on the ``plan-then-execute" approach, improves the performance of GPT-4o by 1.1 points, placing it first among all models. However, CoT is not a one-size-fits-all solution. When LLaVA-v1.6-34B used CoT, its performance dropped by 3.2 points, indicating that the improvement brought by CoT has prerequisites and depends on the model's inherent multi-step reasoning capability. Recently, several CoT methods tailored for VLMs have emerged, such as DCoT~\citep{wu2023despcot} and CCoT~\citep{mitra2024ccot}, but experimental results indicate that their performance remains suboptimal. We attribute this to UAV images that contain numerous small objects, making it difficult to fully describe or relate them compared to natural images. 

Moreover, we identify that the performance degradation of LLaVA-v1.6-34B with CoT primarily stems from its score drop in statistics-type questions (from 29.6 to 22.1). We examined the 61 questions that were answered correctly without CoT but incorrectly with CoT, and found that the model tended to output "Not possible" or "Unsolvable" when using CoT. Statistical analysis revealed 46 such cases (75.4\%), indicating that while the model engaged in reasoning with CoT, its lack of statistical knowledge prevented it from reaching correct conclusions.

\section{Conclusion}

In this work, we propose \dataset, a novel benchmark designed to evaluate the mathematical reasoning capabilities of VLMs in the context of aerial vehicle imagery. We evaluated 14 prominent models and observed that even advanced models like GPT-4o struggle due to a lack of domain-specific knowledge. Based on our experimental results, we conducted an in-depth analysis of the limitations of existing VLMs, offering valuable insights for future research. Additionally, we explored leveraging parameter-efficient fine-tuning to incorporate domain-specific knowledge and Chain-of-Thought prompting to enhance the model’s reasoning abilities. Both methods can effectively improve the performance of VLMs on our benchmark.  \dataset and \textsc{AVI-Math-215k} provide a valuable foundation for future advancements in reliable autonomous aerial systems.











\printcredits

\bibliographystyle{cas-model2-names}

\bibliography{cas-refs}



\end{document}